\documentclass[lettersize,journal]{IEEEtran}
\usepackage{amsmath,amsfonts}
\usepackage{algorithmic}
\usepackage{algorithm}
\usepackage{array}
\usepackage[caption=false,font=normalsize,labelfont=sf,textfont=sf]{subfig}
\usepackage{textcomp}
\usepackage{stfloats}
\usepackage{url}
\usepackage{verbatim}
\usepackage{graphicx}
\usepackage{cite}
\usepackage{xcolor}
\usepackage{amssymb}
\usepackage{booktabs}
\usepackage{color}
\usepackage{colortbl}
\usepackage{bbm}
\usepackage{multirow}
\usepackage{cite}
\definecolor{origin}{rgb}{0.2,0.3,0.9}
\definecolor{mygray}{gray}{.9}
\definecolor{mygray2}{gray}{.8}
\begin{document}

\title{Anchor3DLane++: 3D Lane Detection via Sample-Adaptive Sparse 3D Anchor Regression}

\author{Shaofei Huang, Zhenwei Shen, Zehao Huang, Yue Liao, Jizhong Han, Naiyan Wang, Si Liu
    % <-this % stops a space
\thanks{Shaofei Huang and Jizhong Han are with Institute of Information Engineering, Chinese Academy of Sciences, Beijing, China, and also with School of Cyber Security, University of Chinese Academy of Sciences, Beijing, China. Email: nowherespyfly@gmail.com, hanjizhong@iie.ac.cn.}
\thanks{Zhenwei Shen, Zehao Huang, and Naiyan Wang are with TuSimple, Beijing, China. Email: shenzhenwei@outlook.com, \{zehaohuang18, winsty\}@gmail.com.}
\thanks{Yue Liao is with The Chinese University of Hong Kong, Hong Kong SAR, China, and also with The Chinese University of Hong Kong, Shenzhen, China. Email: liaoyue.ai@gmail.com.}
\thanks{Si Liu is with Institute of Artificial Intelligence, Beihang University, Beijing, China, and also with Hangzhou Innovation Institute, Beihang University, Hangzhou, China. Email: liusi@buaa.edu.cn.}
\thanks{The corresponding author is Si Liu.}
\thanks{A preliminary version of this research has appeared in CVPR 2023~\cite{anchor3dlane}.}}

\markboth{IEEE Transactions on Pattern Analysis and Machine Intelligence}%
{Shell \MakeLowercase{\textit{et al.}}: A Sample Article Using IEEEtran.cls for IEEE Journals}

\maketitle

\begin{abstract}
In this paper, we focus on the challenging task of monocular 3D lane detection.
Previous methods typically adopt inverse perspective mapping (IPM) to transform the Front-Viewed (FV) images or features into the Bird-Eye-Viewed (BEV) space for lane detection.
However, IPM's dependence on flat ground assumption and context information loss in BEV representations lead to inaccurate 3D information estimation.
Though efforts have been made to bypass BEV and directly predict 3D lanes from FV representations, their performances still fall behind BEV-based methods due to a lack of structured modeling of 3D lanes.
In this paper, we propose a novel BEV-free method named Anchor3DLane++ which defines 3D lane anchors as structural representations and makes predictions directly from FV features.
We also design a Prototype-based Adaptive Anchor Generation (PAAG) module to generate sample-adaptive sparse 3D anchors dynamically.
In addition, an Equal-Width (EW) loss is developed to leverage the parallel property of lanes for regularization.
Furthermore, camera-LiDAR fusion is also explored based on Anchor3DLane++ to leverage complementary information.
Extensive experiments on three popular 3D lane detection benchmarks show that our Anchor3DLane++ outperforms previous state-of-the-art methods. 
Code is available at: \url{https://github.com/tusen-ai/Anchor3DLane}.
\end{abstract}

\begin{IEEEkeywords}
Autonomous Driving, Monocular 3D Lane Detection, Sample-Adaptive Sparse Anchors, Anchor Regression
\end{IEEEkeywords}

\section{Introduction}
\begin{figure}[!t]
    \centering
    \includegraphics[width=\linewidth]{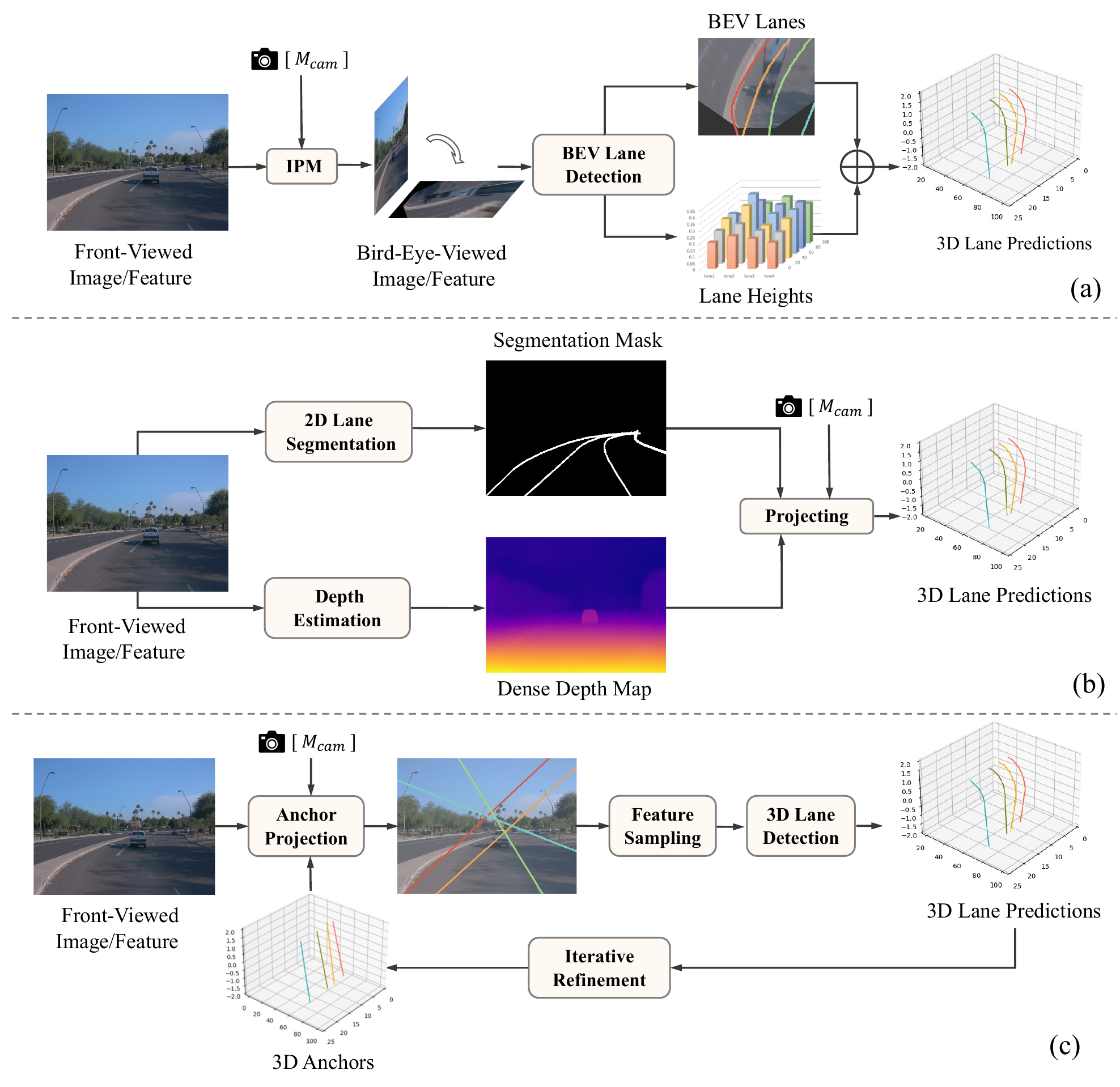}
    \caption{(a) BEV-based methods, which perform lane detection in the warped BEV images or features. (b) Non-BEV method, which projects 2D lane predictions back to 3D space with estimated depth. (c) Our Anchor3DLane++ projects 3D anchors into FV space to sample features for direct 3D prediction.}
    \label{fig:intro}
\end{figure}
Autonomous driving~\cite{li2022bevformer,wu2023heightformer,qin2022uniformer,qin2022monoground,zhou2021monoef} has drawn remarkable attention from researchers in the fields of both academia and industry.
3D lane detection, which involves the identification and localization of lane lines in the 3D space, plays a critical role in the realm of autonomous driving systems.
Accurate and robust perception of 3D lanes is pivotal for ensuring the safety and reliability of self-driving vehicles, not only facilitating lane-keeping but also supporting downstream tasks such as high-definition map construction~\cite{li2023lanesegnet,liu2020high,wang2024openlane}, and trajectory planning~\cite{zhu2020trajectory,jia2023hdgt,wu2022trajectory}.
In this paper, we focus primarily on the task of monocular 3D lane detection where 3D lanes are estimated directly from a frontal-viewed image, which is very challenging due to the lack of accurate depth information.

Current mainstream 3D lane detection methods~\cite{3dlanenet,genlanenet,clgo,persformer} typically detect lanes in the Bird's Eye View (BEV) space, where 3D lanes have more structured geometric properties and better scale consistency. 
A common practice of these BEV-based methods is illustrated in Fig.~\ref{fig:intro}(a).
Images or feature maps captured from the frontal view (FV) are first warped into BEV using inverse perspective mapping (IPM), converting the challenging 3D lane detection problem into 2D lane detection within the BEV space.
The coordinates of the sampled lane points in BEV are combined with their corresponding estimated height values, provided by a height estimation head, to re-project the lanes back into 3D space.
While the above methods have demonstrated good performances, they still have limitations.
First, IPM depends heavily on the flat ground assumption, which becomes invalid for uphill or downhill scenarios, resulting in misaligned 3D coordinate estimations of lane lines under such road conditions.
Second, IPM performs the perspective transformation based on the ground plane, which inevitably loses valuable clues like height and context information above the road's surface.
In addition, objects on the road such as vehicles may undergo severe distortion after IPM, causing the lane markings to be occluded.

The above limitations hinder the accuracy of 3D information restoration from BEV representations, motivating efforts to detect 3D lanes from the FV representations directly~\cite{once,curveformer}.
A typical example~\cite{once} in Fig.~\ref{fig:intro}(b) reconstructs 3D lanes through the combination of 2D lane segmentation and dense depth estimation results.
The 3D locations of lanes can be determined by utilizing estimated depth values and camera parameters to project their 2D segmentation masks into 3D space.
Although these methods circumvent the weaknesses of BEV representations, their performances remain inferior to state-of-the-art BEV-based methods due to inadequate structured modeling of 3D lane lines.

In this paper, we introduce a novel BEV-free method named Anchor3DLane++, which allows for the direct prediction of 3D lanes from FV representations.
As shown in Fig.~\ref{fig:intro}(c), 3D lane anchors are defined as rays in the 3D space with given pitches and yaws in our Anchor3DLane++.
We project points on the 3D anchors onto the FV feature map using camera parameters and then obtain the corresponding anchor features by sampling in the neighborhood of the projected 2D points.
Based on the sampled anchor features, classification and regression results for each anchor can be generated to predict 3D lanes directly.
In the above feature extraction and lane prediction processes, our 3D lane anchors serve as intermediaries that bridge the gap between FV and 3D spaces, thereby facilitating direct prediction from FV representations.
Compared to the information loss in BEV representations, extracting features directly from the original FV features also preserves more contextual information, which in turn improves the precision of 3D lane detection.
Moreover, based on the design of 3D lane anchors, we can conveniently carry out multiple stages of iterative refinement of lane predictions to further boost detection performances.

In Anchor3DLane~\cite{anchor3dlane} (our conference version), to sufficiently cover different positions and shapes of lane lines under diverse road conditions, we place dense anchors at varying angles in the 3D space for lane regression.
However, dense 3D anchor regression has certain limitations.
On the one hand, dense anchors rely heavily on heuristic designs of anchor hyper-parameters and label assignment, which leads to laborious hyper-parameter tuning for different datasets.
On the other hand, performing feature sampling and prediction for an excessive number of 3D anchors results in significant computational redundancy, given that real-world road scenarios typically feature only a limited number of visible lane lines simultaneously.
Therefore, we investigate sparse 3D anchor regression in our Anchor3DLane++ framework to alleviate the limitations derived from dense anchors.
To avoid performance degradation resulting from anchor sparsification, we propose a novel Prototype-based Adaptive Anchor Generation (PAAG) module that learns anchor prototypes from datasets and dynamically combines them to produce sample-adaptive sparse anchors.
Through the adaptive combination of prototypes, it is possible to cover potential positions and shapes of lanes per image sample using a small number of well-aligned anchors, thus yielding more accurate 3D lane predictions.

In addition, we also explore the effects of camera-LiDAR fusion based on our Anchor3DLane++ framework.
The heterogeneous nature of camera and LiDAR modalities, where cameras capture dense grid data and LiDAR captures sparse point data, makes their alignment and fusion a challenging task.
Thanks to the anchor projection and feature sampling in Anchor3DLane++, we can easily acquire spatially aligned features of different modalities associated with the same 3D anchor, which are more amenable to fusion. 
By integrating the anchor features from camera data containing rich texture information and LiDAR data containing precise depth information, a complementary effect is achieved to enhance both the accuracy and reliability of 3D lane detection.

Moreover, we also degign an Equal-Width (EW) loss that constrains the widths of 3D lane proposals to be consistent.
The motivation is based on an intuitive observation that lanes usually appear parallel on the same road surface except for the fork lanes.
Therefore, the width between each pair of non-fork lane lines tends to be nearly consistent when measured at different locations.
By applying this geometric property as a regularization, our Anchor3DLane++ is optimized within a narrowed solution space, which mitigates the inherent ill-posed issue of monocular 3D lane detection and obtains more accurate and robust detection results.

Our contributions are summarized as follows:
(1) We introduce a novel Anchor3DLane++ framework, which defines lane anchors in 3D space and directly detects 3D lanes from FV representations, eliminating the reliance on BEV space.
A simple yet effective camera-LiDAR fusion approach is also explored upon the Anchor3DLane++ framework for enhanced 3D lane detection.
(2) We propose a novel Prototype-based Adaptive Anchor Generation (PAAG) module that dynamically combines anchor prototypes to produce sample-adaptive sparse anchors, thus making sparse anchor regression feasible for 3D lanes.
(3) We design an Equal-Width (EW) loss to leverage the parallel property of lane lines for model regularization.
(4) Extensive experimental results on three popular 3D lane detection benchmarks demonstrate that our Anchor3DLane++ outperforms previous state-of-the-art methods.

This paper is built upon Anchor3DLane~\cite{anchor3dlane} (our conference version) and significantly extends it in several aspects.
First, we introduce an improved framework named Anchor3DLane++, which replaces the original paradigm of dense anchor regression with sample-adaptive sparse anchor regression, thus alleviating the limitations of heuristic design and redundant computation of dense anchors.
A novel Prototype-based Adaptive Anchor Generation (PAAG) module is proposed in Anchor3DLane++ to dynamically produce sparse anchors aligned with the input images, which avoids performance degradation of sparse anchors and yields more accurate predictions.
Second, we design a new Equal-Width (EW) loss that modifies the original offline equal-width constraint into an online regularization term, resulting in improved performance and reduced time costs.
Third, based on our Anchor3DLane++ framework, we explore the fusion of camera and LiDAR modalities to leverage their complementarity for further performance boost.
Fourth, we supplement with a substantial amount of new experimental results, including comparison with previous methods, ablation studies, qualitative analysis, \textit{etc}.
Compared to the conference version, our extended method achieves large performance gains (\textit{e.g.}, $+9.2$\% on OpenLane~\cite{persformer} dataset for F1 score).

\section{Related Works}
\subsection{2D Lane Detection}
2D lane detection~\cite{qin2020ultra,wang2022keypoint,linecnn,liu2021end,tabelini2021polylanenet,pan2018spatial,jin2022eigenlanes,yang2023lane} focuses on accurately delineating the shape and position of 2D lane lines in the given image and distinguishing between different instances of them as well.
Traditional works~\cite{aly2008real, zhou2010novel, kim2008robust, wang2004lane} predominantly concentrated on the extraction of low-level hand-crafted features, including edges and textures.
These methods, however, often involve intricate feature extraction and complex post-processing procedures, thus exhibiting a lack of robustness in ever-changing diverse environments.
With the development of deep learning, approaches based on neural networks have been widely investigated and demonstrated significant performance improvements. Segmentation-based frameworks~\cite{pan2018spatial, hou2019learning, neven2018towards,qiu2023priorlane} formulate 2D lane detection as a semantic/instance segmentation task and the key to these methods is the development of more efficient and semantically rich feature extraction techniques.
To make predictions more sparse and flexible, keypoint-based methods~\cite{qu2021focus, wang2022keypoint, ko2021key, rclane} model lane lines as sequences of ordered keypoints and utilize post-processing to associate keypoints belonging to the same lane.
Different from these bottom-up methods above, detection-based methods~\cite{linecnn, laneatt, condlanenet, clrnet, chen2023generating,han2022laneformer,condlanenet,qin2020ultra,qin2022ultra} detect lane lines following a top-down manner, where predictions are made by defining lane anchors and regressing offsets from anchor points to corresponding lane points.
Non-Maximum Suppression (NMS) is typically applied in these methods to keep lanes with higher confidence.

\subsection{3D Lane Detection}
Due to the reduced accuracy and robustness in projecting 2D lanes into 3D space, there is an increasing exploration for the task of 3D lane detection~\cite{bai2018deep,3dlanenet} which directly identifies and localizes lanes in 3D space.
Since cameras are widely used in autonomous driving systems, monocular 3D lane detection~\cite{3dlanenet, genlanenet, clgo, once, 3dlanenet+, ai2023ws} receives much attention.
Among these methods, owing to the superior geometric properties of 3D lanes when viewed from a BEV perspective, a prevalent solution for 3D lane detection is transforming FV representations to BEV and making predictions in the BEV space. 
For example, 3DLaneNet~\cite{3dlanenet} employs IPM to transform FV features to BEV, followed by anchor offsets regression of BEV lanes. 
CLGo~\cite{clgo} warps the raw images to BEV images utilizing estimated camera poses to get rid of the reliance on camera calibrations.
Given that IPM predominantly depends on the assumption of flat ground, predicting lanes in BEV space may result in misalignment with the real 3D space in scenarios involving uneven terrain. 
Gen-LaneNet~\cite{genlanenet} further resolves the spatial alignment by distinguishing between the virtual top view produced by IPM and the real top view in 3D space. 
PersFormer~\cite{persformer} leverages deformable attention mechanisms to produce BEV features more robustly.
M$^2$-3DLaneNet~\cite{m2net} proposes a multi-modal fusion method that utilizes LiDAR points to lift image features into 3D space and fuse camera-LiDAR features in the BEV space.
Apart from these methods specially designed for 3D lane detection, BEV representations are also widely adopted in HD map construction, such as MapTR~\cite{MapTR} and MapTRv2~\cite{maptrv2}.
These methods model map elements such as lane lines and pedestrian crossings by map queries and predict them as a sequence of points to form a polyline or polygon based on BEV features.
StreamMapNet~\cite{yuan2024streammapnet} further employs a streaming strategy to propagate history BEV features and map queries for long-range temporal information fusion, achieving significant performance improvement.

To further cast off BEV representations, SALAD~\cite{once} first decomposes 3D lane detection into two subtasks, namely 2D lane segmentation and dense depth estimation. 
However, the absence of structured modeling for 3D lanes limits its performance.
Concurrent to our Anchor3DLane~\cite{anchor3dlane}, CurveFormer~\cite{curveformer} and LATR~\cite{luo2023latr} adopt a DETR~\cite{detr}-like Transformer architecture and defines lane queries to extract query embeddings directly from FV features for lane prediction, which is still implicit in 3D lane modeling.
Besides, LATR follows a two-stage pipeline that first conducts 2D image segmentation to obtain initial query embeddings and then decodes lane lines in the Transformer decoder, which is not as straightforward as our one-stage pipeline.
Unlike the above methods, our Anchor3DLane~\cite{anchor3dlane} and Anchor3DLane++ define anchors in the 3D space for explicit 3D lane modeling, thereby facilitating the bridging between FV space and 3D space. 
Our anchor projection and feature sampling designs ensure accurate anchor feature extraction, which enables the precise prediction of 3D lanes directly from FV representations and circumvents the introduction of BEV.

\subsection{Sparse Object Formulation}
Unlike dense object modeling methods~\cite{focalloss,fcos} which utilize dense predefined object candidates to make predictions across all positions, sparse object modeling methods typically define a fixed number of sparse object priors and allow the model to focus on limited high-quality candidates.
For example, DETR~\cite{detr} introduces a query-based object detection approach, utilizing a set of learnable object queries to reason about the relationships between objects and the global context through multiple Transformer decoder layers.
However, it suffers from slow convergence due to the unconstrained object queries.
To tackle this issue, its followers~\cite{zhu2020deformable,wang2022anchor} introduce spatial priors into object queries to provide spatial information, thereby accelerating the convergence process.
Sparse R-CNN~\cite{sun2023sparse} and its subsequent works~\cite{hong2022dynamic,zhang2022featurized} introduce a sparse set of learnable or image-dependent proposal boxes and features, which are then utilized to extract RoI features for regression and classification explicitly.

In the domain of 3D lane detection, existing sparse detectors~\cite{clgo,curveformer} predominantly adopt a query-based detection scheme, which involves using a set of learnable lane queries to interact with image context through attention mechanisms for feature aggregation and predicts polynomial coefficients of each lane line.
However, due to the slender shape of lane lines, the naive imitation of query-based object detection paradigms results in sub-optimal lane modeling and prediction. 
Besides, lane queries lack explicit priors for lane shapes and positions, leading to relatively low convergence speed and inferior prediction accuracy.
Distinct from the above methods, our Anchor3DLane++ employs sample-adaptive sparse 3D lane anchors that contain explicit shape and position priors to model 3D lanes in a structured manner, thus achieving more accurate lane feature extraction and coordinate regression.

\section{Anchor3DLane++}
\label{sec:method}

\begin{figure*}[!t]
    \centering
    \includegraphics[width=\linewidth]{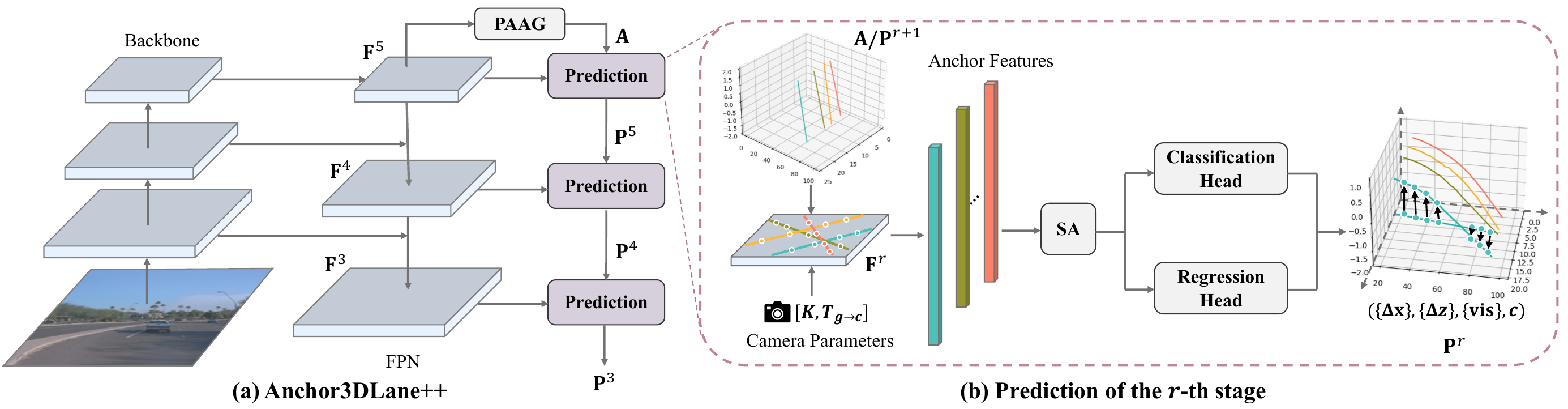}
    \caption{The overall architecture of Anchor3DLane++. (a) Pipeline of Anchor3DLane++. Proposals output from the previous stage are used as the new anchors for the next stage. (b) The $r$-th iterative stage of anchor projection and 3D lane prediction. 3D anchors or 3D proposals from $(r+1)$-th stage are projected to sample their features from $\mathbf{F}^r$ using camera parameters. A classification head and a regression head are applied after a self-attention layer to make prediction.}
    \label{fig:pipeline}
\end{figure*}

We present the overall architecture of our Anchor3DLane++ in Fig.~\ref{fig:pipeline}(a).
An input front-viewed image $\mathbf{I}\in \mathbb{R}^{H_I \times W_I \times 3}$ is processed by a CNN (ResNet~\cite{resnet} in our paper) backbone and an FPN~\cite{fpn} neck sequentially to extract multi-level 2D FV visual features, where $H_I$ and $W_I$ denote the height and width of the image respectively.
The extracted features are denoted as $\mathbf{F}^r \in \mathbb{R}^{H_F \times W_F \times C_F}$, where $r \in \{3, 4, 5\}$ corresponds to the stage number of ResNet, and $H_F = H_I / 8$, $W_F = W_I / 8$, and $C_F$ represent the height, width and channel number of the feature maps respectively.
$\mathbf{F}^5$ is first fed into the Prototype-based Adaptive Anchor Generation (PAAG) module to obtain initial sparse 3D anchors $\mathbf{A}$, which offer essential positional priors contextualized by the input image.
Predictions are made based on the features of $\mathbf{A}$ sampled from $\mathbf{F}^5$, and then utilized as the new 3D anchors in the next stage of iterative refinement.
Predictions from the last stage are taken as the final outputs of Anchor3DLane++.

\subsection{Preliminaries}
\begin{figure}[!htbp]
    \centering
    \includegraphics[width=\linewidth]{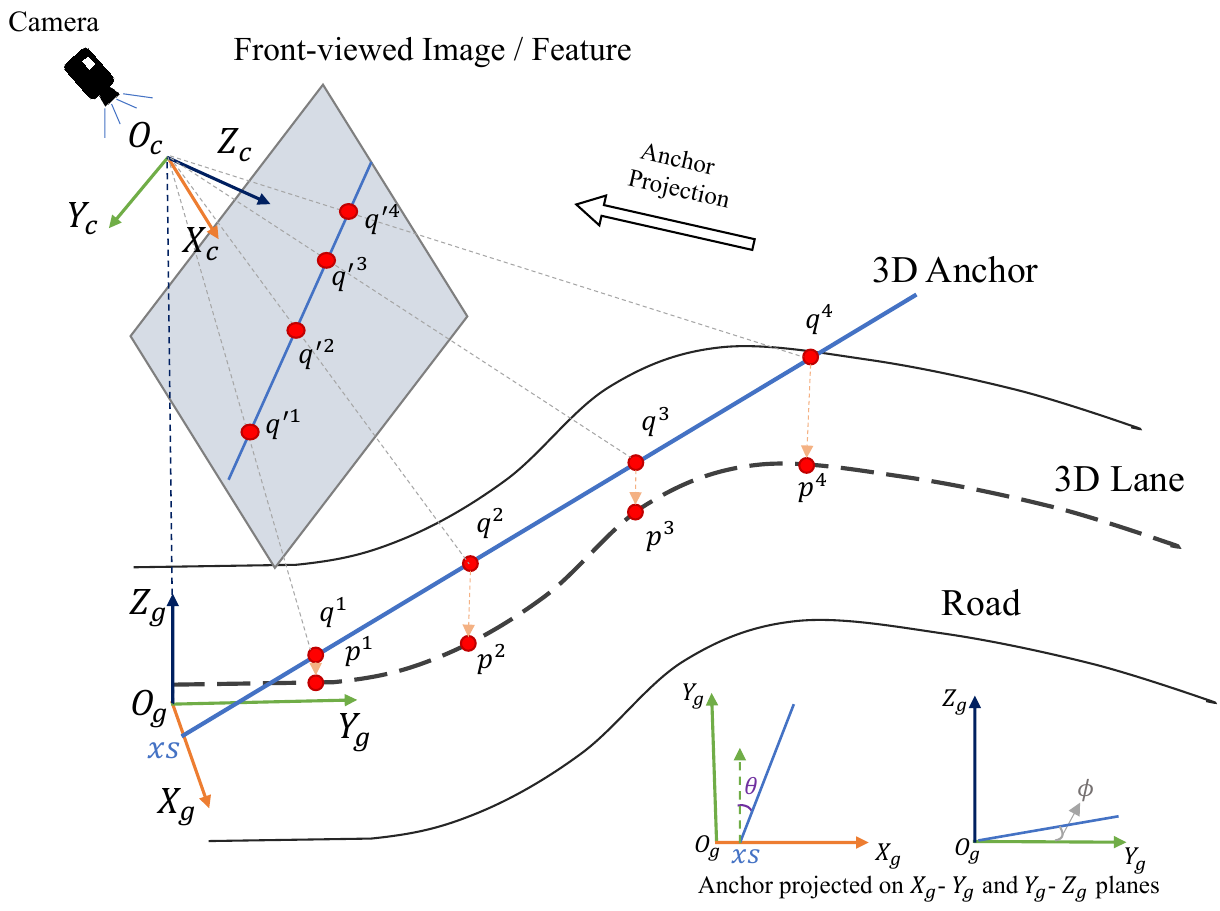}
    \caption{Illustration of 3D anchor and 3D lane in the ground coordinate system.}
    \label{fig:anchor}
\end{figure}

We first review the representation of coordinate systems and 3D lanes.
As shown in Fig.~\ref{fig:anchor}, 3D lanes are typically annotated in the ground coordinate system, which is defined by origin $O_g$ and $X_g$, $Y_g$, and $Z_g$ axes.
Specifically, $O_g$ is positioned on the road directly below the camera center, x-axis $X_g$ points positively to the right, y-axis $Y_g$ points positively forward and z-axis $Z_g$ points positively upward.
Within the ground coordinate system, each 3D lane is represented as a sequence comprising $N$ 3D points, whose y-coordinates $\mathbf{y}=\{y^k\}_{k=1}^N$ are uniformly sampled along $Y_g$.
Taking the $i$-th 3D lane $\mathbf{G}_i=\{\mathbf{p}_i^k\}_{k=1}^N$ as an example, its $k$-th point is described as $\mathbf{p}_i^k=(\hat{x}_i^k, y^k, \hat{z}_i^k, \hat{vis}_i^k)$, where the first three elements indicate the coordinates of $\mathbf{p}_i^k$, and the last element $\hat{vis}_i^k \in \{0, 1\}$ indicates its visibility in case the lane is partially visible in the view.
Given a camera mounted on the ego-vehicle, the camera coordinate system which aligns with the front-view (FV) image, is a right-handed system defined by its origin $O_c$, and axes $X_c, Y_c, Z_c$, wherein $O_c$ located at the center of the camera and $Z_c$ extending forward perpendicular to the camera plane.
Through the well-calibrated camera parameters, a 3D point in the ground coordinate system can be projected to the space of the captured FV image or image feature, which is elaborated in Sec.~\ref{sec:sample}.
In line with common practices in prior works~\cite{3dlanenet, genlanenet}, our Anchor3DLane++ primarily uses the ground coordinate system for 3D representation.
However, it can easily adapt to other 3D coordinate systems, provided that camera calibration parameters are accessible.

\subsection{Prototype-based Adaptive Anchor Generation}
In our method, we define 3D lane anchors within the same coordinate systems as the 3D lanes, \textit{i.e.}, the ground coordinate, for easier coordinate regression.
As illustrated in Fig.~\ref{fig:anchor}, a 3D anchor is defined as a ray emitted from the $X_g$ axis.
Each ray is uniquely determined by its starting coordinates $(xs, 0, 0)$, the angle $\phi$ between its projection on the $X_g$-$Y_g$ plane and the $Y_g$ axis, and the angle $\theta$ between its projection on the $Y_g$-$Z_g$ plane and the $Y_g$ axis.
These three variables are collectively referred to as the \textbf{Anchor Metas} in this paper.
Similar to 3D lane, the 3D anchor is also represented by $N$ 3D points sampled with the same y-coordinates as 3D lanes, and the $j$-th 3D anchor can be denoted as $\mathbf{A}_j=\{\mathbf{q}_j^k\}_{k=1}^N$, with its $k$-th point being $\mathbf{q}_j^k=(x_j^k, y^k, z_j^k)$.
In Anchor3DLane~\cite{anchor3dlane}, to cover as many lane positions and shapes as possible, we exhaustively enumerate all potential values of anchor metas and combine them using the Cartesian Product, thus resulting in a cubic order of magnitude of dense anchors.
However, dense anchors introduce drawbacks such as heuristic design and redundant computations.
Therefore, we propose a novel Prototype-based Adaptive Anchor Generation (PAAG) module in Anchor3DLane++ to generate sample-adaptive sparse anchors based on meta prototypes, thus sparsifying 3D anchors while maintaining the coverage of lane lines simultaneously.

\begin{figure}[!htbp]
    \centering
    \includegraphics[width=\linewidth]{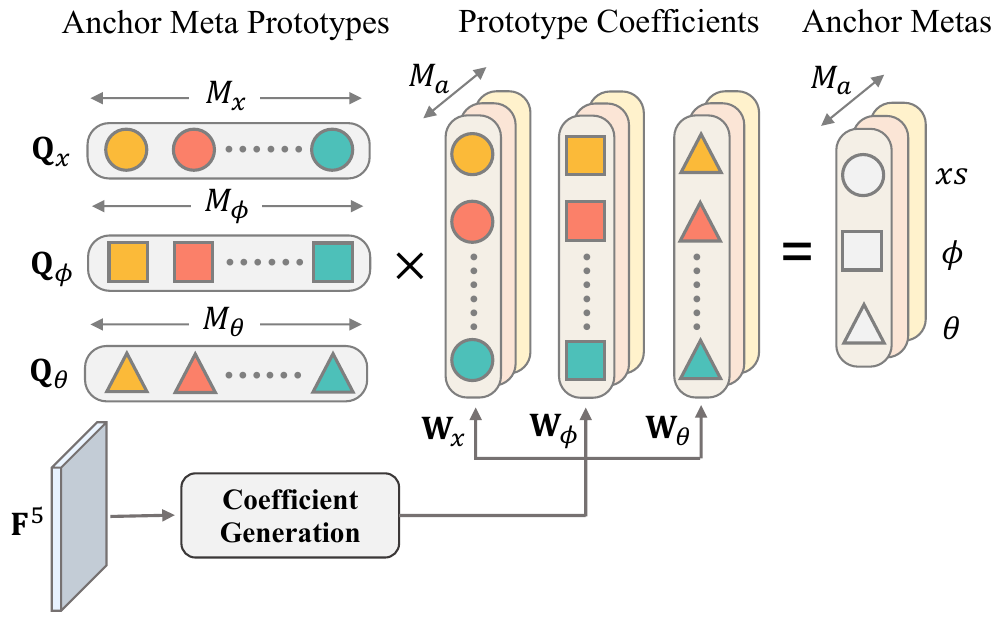}
    \caption{Illustration of Prototype-based Adaptive Anchor Generation.}
    \label{fig:proto}
\end{figure}

The details of the PAAG module are shown in Fig.~\ref{fig:proto}.
First, we define a set of learnable parameters for each anchor meta as its prototypes, and use $\mathbf{Q}_x\in \mathbb{R}^{M_x}$, $\mathbf{Q}_\phi \in \mathbb{R}^{M_\phi}$, and $\mathbf{Q}_\theta \in \mathbb{R}^{M_\theta}$ to denote the prototypes for starting coordinate $xs$, angle $\phi$, and angle $\theta$ respectively, where $M_x$, $M_\phi$, and $M_\theta$ represent the number of prototypes for each anchor meta.
To facilitate optimization, all meta prototype elements are set between $[-1, 1]$ through min-max normalization initially.
Through continuous updates on the training set, these meta prototypes progressively learn the potential shapes and starting positions of lane lines, enabling the generation of diverse anchor metas through linear combinations of these prototypes.
To generate anchors that align better with the lane lines in the input image, we utilize the visual feature $\mathbf{F}^5$ to obtain the prototype coefficients.
Concretely, we first average the values on the height dimension of $\mathbf{F}^5$ and flatten it into a vector.
Afterwards, the visual feature is processed by three distinct linear layers to obtain the coefficients corresponding to the $xs$, $\phi$, and $\theta$ prototypes, denoted as $\mathbf{W}_x\in \mathbb{R}^{M_a\times M_x}$, $\mathbf{W}_\phi \in \mathbb{R}^{M_a\times M_\phi}$, and $\mathbf{W}_\theta \in \mathbb{R}^{M_a \times M_\theta}$.
We use $M_a$ to represent the number of sparse anchors adopted in our Anchor3DLane++.
We apply Softmax function on the second dimension of the coefficients matrices for normalization.
Finally, the dot product is conducted between prototypes and coefficients of each anchor to obtain its anchor metas:
\begin{equation}
xs_j = f(\mathbf{Q}_x\mathbf{W}_{x,j}),~~
\phi_j = f(\mathbf{Q}_\phi\mathbf{W}_{\phi,j}), ~~
\theta_j = f(\mathbf{Q}_\theta\mathbf{W}_{\theta,j}),
\end{equation}
where $j \in [1, M_a]$, and $f(\cdot)$ involves both truncation and scaling operations. 
An example of applying $f(\cdot)$ on $xs_j$ is presented as follows, and $\phi_j$ and $\theta_j$ are processed similarly:
\begin{gather}
\bar{xs}_j = {\rm clip}(\mathbf{Q}_x\mathbf{W}_{x,j}, -1, 1), \\
xs_j = \bar{xs}_j \cdot (xs_{max} - xs_{min}) + xs_{min},
\end{gather}
where ${\rm clip}(\cdot, b_l, b_u)$ denotes the truncation operation between $b_l$ and $b_u$, and $xs_{min}$ and $xs_{max}$ are predefined values of $xs$ ranges.
The initial sparse 3D anchors $\{\mathbf{A}_j\}_{j=1}^{M_a}$ are then generated using the predetermined y-coordinates.

Our PAAG module exhibits advantages in two primary respects. 
On the one hand, dynamically combining different metas to generate anchors eliminates the need to exhaustively enumerate all possible meta combinations.
In this way, it is possible to cover all potential positions and shapes within the image using a small number of anchors, thereby achieving effective anchor sparsification.
On the other hand, by weighing the meta prototypes learned on the training set during inference, we can generate sample-adaptive sparse anchors rather than fixed ones, which adapt to the distribution of testing sets with more flexibility.

\subsection{Anchor Projection and Lane Prediction}
\label{sec:sample}
After generating the sparse 3D anchors, we project them into the plane of an FV feature $\mathbf{F}^r$ at a certain level with the assistance of camera parameters to obtain corresponding anchor features.
The feature index $r$ will be omitted in this section for ease of illustration.
Considering an anchor $\mathbf{A}_j$, we take its $k$-th point $\mathbf{q}_j^k$ as an example to illustrate the projection operation.
The subscript $j$ will be omitted for the sake of simplicity in the following formulations:
\begin{gather}
\label{eq:proj}
    \begin{bmatrix}
    \tilde{u}^k \\
    \tilde{v}^k \\
    d^k \\
    \end{bmatrix} = \mathbf{K} \mathbf{T}_{g \rightarrow c} \begin{bmatrix}
    x^k \\
    y^k \\
    z^k \\
    1
    \end{bmatrix}, \\
    u^k = W_F / W_I \cdot \frac{\tilde{u}^k}{d^k}, \\
    v^k = H_F / H_I \cdot \frac{\tilde{v}^k}{d^k},
\end{gather}
where $\mathbf{K} \in \mathbb{R}^{3\times 3}$ represents camera intrinsic parameters, $\mathbf{T}_{g \rightarrow c} \in \mathbb{R}^{3\times 4}$ represents the transform matrix from ground coordinate to camera coordinate, and $d^k$ denotes the depth of point $\mathbf{q}^k$ to the camera plane.
Through the above formulations, $\mathbf{q}^k$ is projected to point $\mathbf{q}'^k=(u^k, v^k)$ in the space of FV feature $\mathbf{F}$ and its feature $\mathbf{F}_{(u^k, v^k)}$ can be obtained through bilinear interpolation within the neighbourhood of $\mathbf{q}'^k$ on $\mathbf{F}$.
Finally, we obtain the feature of anchor $\mathbf{A}_j$ by concatenating the features of all its projected points along the channel dimension, resulting in the composite anchor feature $\{\mathbf{F}_{(u^k, v^k)}\}_{k=1}^N$.

The obtained anchor features are first processed by a self-attention layer for feature enhancement. For the sampled feature of each anchor, we apply a classification head to predict the classification probabilities $\mathbf{c}_j \in \mathbb{R}^S$ where $S$ represents the total number of lane types, and a regression head to estimate the coordinate offsets $(\Delta \mathbf{x}_j \in \mathbb{R}^N, \Delta \mathbf{z}_j \in \mathbb{R}^N) = \{(\Delta x_j^k, \Delta z_j^k)\}_{k=1}^N$ for all points belonging to this anchor, as well as their visibility scores $\mathbf{vis}_j \in \mathbb{R}^N$. 
In this way, the 3D lane proposals can be obtained as $\mathbf{P}_j=(\mathbf{c}_j, \mathbf{x}_j+\Delta \mathbf{x}_j, \mathbf{y}, \mathbf{z}_j+\Delta \mathbf{z}_j, \mathbf{vis}_j)$, $j \in [1, M_a]$.

To further improve performances, we implement cross-layer iterative refinement~\cite{clrnet} on the multi-level feature pyramid, capitalizing on the distinct characteristics exhibited by features at different levels to achieve a more holistic prediction.
As shown in Fig.~\ref{fig:pipeline}(b), since high-level features contain richer semantic information, we first harness $\mathbf{F}^5$ to obtain the meta prototype coefficients, thus yielding the initial set of sparse anchors $\mathbf{A}$ for the whole iterative procedures.
Anchor features are then sampled from $\mathbf{F}^5$ for prediction as discussed above, resulting in the 3D lane proposals for the $5$-th level ($\mathbf{P}^5$), which serve as new 3D anchors for the next stage of iteration on feature $\mathbf{F}^4$.
This process iterates over the feature pyramid from higher-level features to lower-level ones, and the final prediction is made on the lowest-level feature, utilizing its local context information to achieve more precise coordinate regression.
The 3D anchors progressively conform more closely to the lanes in the image through layer-by-layer refinement, thereby achieving more accurate anchor feature sampling and coordinate regression.

\subsection{Equal-Width Loss}
\label{sec:ewc}

As structured objects, lane lines exhibit the superior geometric property of being nearly parallel to each other on the same road surface in most cases, which could be exploited to impose constraints on the predicted 3D lanes to alleviate the ill-posedness of monocular 3D coordinate estimation.
To this end, we formulate this parallel property of lanes as an Equal-Width (EW) loss which constrains the width between each pair of lane proposals to remain consistent when measured at different sampling points.
Given two lane proposals $\mathbf{P}_{j}=\{\mathbf{p}_j^k\}_{k=1}^N$ and $\mathbf{P}_{j'}=\{\mathbf{p}_{j'}^k\}_{k=1}^N$, 
width between $\mathbf{P}_{j}$ and $\mathbf{P}_{j'}$ measured at point pair $\mathbf{p}_{j}^k$ and $\mathbf{p}_{j'}^k$ can be approximated by assuming lane segment between $\mathbf{p}_{j}^k$ and $\mathbf{p}_{j}^{k+1}$ to be straight:
\begin{equation}
w_{j,j'}^k \approx \vert \cos{\varphi_{j'}^k}(x_{j'}^k - x_j^k) \vert,
\end{equation}
where $\varphi_{j'}^k$ denotes the angle between the normal direction of $\mathbf{P}_{j'}$ at point $\mathbf{p}_{j'}^k$ and x-axis.
The cosine value of $\varphi_{j'}^k$ is calculated as:
\begin{equation}
\cos{\varphi_{j'}^k} = \frac{y^{k+1}-y^k}{\sqrt{(y^{k+1}-y^k)^2+(x_{j'}^{k+1}-x_{j'}^k)^2}}.
\end{equation}
Thus, EW loss for proposal pair ($\mathbf{P}_j,\mathbf{P}_{j'}$) is calculated as:
\begin{equation}
\mathcal{L}_{EW}(j,j') = \begin{cases}
\Delta{w}_{j,j'} & \text{if } \Delta{w}_{j,j'} < \tau \\
0 & \text{otherwise}
\end{cases},
\end{equation}
where $\Delta{w}_{j,j'}$ is the mean absolute deviation of widths:
\begin{equation}
    \Delta{w}_{j,j'}=\frac{1}{N}\sum_{k=1}^N{\vert w_{j, j'}^k-\frac{1}{N}\sum_{k'=1}^N{w_{j, j'}^{k'}}\vert}.
\end{equation}
Due to the existence of special situations such as lane merging or diverging, lane lines might not maintain parallelism at all times.
We set a threshold $\tau$ to exclude these instances, thereby preventing the inappropriate optimization of non-parallel lanes.
Given $M_p$ proposals that are labeled as positive samples (detailed in Sec.~\ref{sec:loss}), the overall EW loss is calculated between all proposal pairs:
\begin{equation}
    \mathcal{L}_{EW} = \frac{1}{M_p(M_p-1)}\sum_{j=1}^{M_p}{\sum_{j'=1, j' \neq j}^{M_p}{\mathcal{L}_{EW}(j,j')}}.
    \label{eq:ewl}
\end{equation}

In our conference version, we design an equal-width constraint to adjust the x-coordinates of the lane predictions in an offline manner.
However, optimization by equal-width constraint requires repeated iteration until convergence, leading to a relatively high time cost.
Besides, once the model parameters are fixed, offline optimization can only make limited adjustments.
In Anchor3DLane++, we modify the offline constraint into a loss function, which can be incorporated as a regularization term of the total loss for end-to-end training.
In this way, the solution space of our model is narrowed down to enhance the accuracy and robustness of lane detection.

\subsection{Overall Loss Functions}
\label{sec:loss}
During training, one-to-one matching is adopted to associate each ground-truth lane with only one proposal as positive using an optimal bipartite matching algorithm.
Given a ground-truth $\mathbf{G}_i$ with class index $s_i$ and a 3D proposal $\mathbf{P}_j$, the matching cost is defined as:
\begin{equation}
\label{eq:cost}
\mathcal{C}(\mathbf{G}_i,\mathbf{P}_j) = -\beta_{cls}\mathbf{c}_j^{s_i} + \beta_{dis}\mathcal{D}(\mathbf{G}_i, \mathbf{P}_j),
\end{equation}
where $\beta_{cls}$ and $\beta_{dis}$ represent the coefficients to balance between classification and distance costs, and $\mathcal{D}(\cdot,\cdot)$ represents the distance metric which is calculated as follows:
\begin{equation}
\label{eq:dis}
    \mathcal{D}(\mathbf{G}_i, \mathbf{P}_j) = \frac{\sum_{k=1}^N{\hat{vis}_i^k \cdot \sqrt{(\hat{x}_i^k - x_j^k)^2 + (\hat{z}_i^k - z_j^k)^2}}}{\sum_{k=1}^N{\hat{vis}_i^k}}.
\end{equation}
After the above matching process, let $s_j$ denote the index of the assigned label (including the non-lane class) for the $j$-th proposal, the classification loss is calculated as:
\begin{equation}
    \mathcal{L}_{cls} = -\sum_{j=1}^{M_a}{\log{\mathbf{c}_j^{s_j}}},
\end{equation}
The regression loss is only calculated between the ground-truth lanes and their assigned positive proposals following~\cite{genlanenet}:
\begin{equation}
\begin{aligned}
    \mathcal{L}_{reg} &= \sum_{i=1}^{M_p}{\Vert \hat{\mathbf{vis}}_i \cdot (\mathbf{x}_{\sigma(i)} + \Delta \mathbf{x}_{\sigma(i)} -\hat{\mathbf{x}}_i) \Vert_1 } \\
    &+ \sum_{i=1}^{M_p}{\Vert \hat{\mathbf{vis}}_i \cdot (\mathbf{z}_{\sigma(i)} + \Delta \mathbf{z}_{\sigma(i)} -\hat{\mathbf{z}}_i) \Vert_1} \\
    &+ \sum_{i=1}^{M_p}{\Vert \mathbf{vis}_{\sigma(i)}-\hat{\mathbf{vis}}_i \Vert_1},
\end{aligned}
\end{equation}
where $\sigma(i)$ is used to denote the index of the proposal assigned to the $i$-th ground-truth lane.

The total loss function of our Anchor3DLane++ is summarized as follows:
\begin{equation}
    \mathcal{L}_{total} = \lambda_{cls} \mathcal{L}_{cls} + \lambda_{reg} \mathcal{L}_{reg} + \lambda_{EW} \mathcal{L}_{EW},
\end{equation}
where $\lambda_{cls}$, $\lambda_{reg}$ and $\lambda_{EW}$ denote the loss coefficients.

\subsection{Camera-LiDAR Fusion}
\label{sec:lc}
Owing to the extensible design of 3D anchors, our Anchor3DLane++ can easily realize camera-LiDAR fusion for more precise and reliable predictions.
Specifically, given LiDAR points corresponding to the FV image, a point encoder (\textit{e.g.}, SECOND~\cite{yan2018second} or PointPillars~\cite{lang2019pointpillars}) is employed to extract multi-stage LiDAR features $\mathbf{F}_L^r \in \mathbb{R}^{D^r \times H_L^r \times W_L^r \times C_L^r}$ for fusion with corresponding stage of image feature, where $r \in \{3,4,5\}$ corresponds to the stage number of the point encoder. Here, $D^r$, $H_L^r$, and $W_L^r$ correspond to the spatial size of the LiDAR feature from the $r$-th stage, and $C_L^r$ represents its channel number. Take feature $\mathbf{F}_L^r$ as an example, we omit the superscript $r$ for simplicity.
For a certain 3D anchor $\mathbf{A}_j$, we project its sampling points $\{\mathbf{p}_j^k\}_{k=1}^N$ onto $\mathbf{F}_L$ using the transformation matrix $\mathbf{T}_{g\rightarrow l} \in \mathbb{R}^{3\times 4}$ to obtain LiDAR features of these points, and the projection operation is realized via:
\begin{equation}
\label{eq:temp_proj}
    \begin{bmatrix}
    x_j^{k'} \\
    y_j^{k'} \\
    z_j^{k'} \\
    \end{bmatrix} = \mathbf{T}_{g\rightarrow l} \begin{bmatrix}
    x_j^k \\
    y_j^k \\
    z_j^k \\
    1 \\
    \end{bmatrix},
\end{equation}
where $x_j^{k'}$, $y_j^{k'}$ and $z_j^{k'}$ represent the coordinates of the projected points.
Thus, the LiDAR feature of $\mathbf{A}_j$ is also acquired through the concatenation of point features in a similar way as its image feature.
We concatenate the features from the two modalities to obtain the multimodal feature of each anchor, which is subsequently fed into the classification head and regression head for proposal prediction.
Leveraging both the rich texture information in the camera images and the precise depth cues in the LiDAR points for complementarity, the performances of 3D lane detection can be further boosted.
\section{Experiments}

\subsection{Experimental Setting}
\subsubsection{Datasets}
Experiments are conducted on three popular 3D lane detection benchmarks,
including OpenLane~\cite{persformer}, ApolloSim~\cite{genlanenet}, and ONCE-3DLanes~\cite{once}.

\textbf{OpenLane} is a large-scale real-world benchmark for 3D lane detection and is built upon the Waymo Open dataset~\cite{waymo}.
It comprises 1000 sequences, which include 200K frames in total, with each frame including both camera intrinsic and extrinsic.
More than 880K lanes from 14 categories are annotated, including those on the opposite side if there is no curbside in the middle of the scene.
Additionally, scene tags (\textit{e.g.}, weather conditions and locations) are offered.
The LiDAR points are obtained from the Waymo Open dataset.
% The LiDAR points corresponding to the annotated frames are all obtained from the original Waymo Open dataset.

\textbf{ApolloSim} is a photo-realistic synthetic dataset that is created using the Unity 3D engine. It comprises over 10.5K images derived from various environments such as highways, urban streets, residential areas, and downtown settings. 
Weather conditions, daytime, traffic/obstacles, and road surface qualities are also diverse.
Three different scenes are included, \textit{i.e.}, balanced scenes, rare subset, and visual variations.

\textbf{ONCE-3DLanes} is a large-scale real-world 3D lane detection dataset.
It comprises 211K annotated frames in total, covering diverse weather conditions (\textit{e.g.}, sunny, cloudy, and rainy) and different environments (\textit{e.g.}, suburbs, bridges, highways, downtown, and tunnels).
Only camera intrinsic for each frame is available in this dataset.

\subsubsection{Evaluation Metrics}
During the evaluation process, a minimum-cost flow algorithm is adopted to match lane predictions and ground truth lanes and the matching cost is calculated as the square root of the sum of the pointwise Euclidean distance between the sampled points.
If more than $75\%$ of the points' distances between a prediction and a ground-truth lane are below $1.5$m, this prediction will be considered as true positive.
Utilizing this definition, the Average Precision (AP) and the maximum F1 score can be calculated.
Besides, x errors and z errors are also counted at both near ($0$-$40$m) and far ($40$-$100$m) ranges.
For the ApolloSim dataset, AP, F1 score, and x/z errors at near and far ranges are utilized as evaluation metrics.
For the OpenLane dataset, in addition to the above ones, category accuracy of all the true positive predictions is also reported.
While for the ONCE-3DLanes, a distinct method for matching predictions and ground truth lanes is employed.
Initially, the matching degree is determined by the IoU between each pair of prediction and ground truth on the top-view plane, where the IoU is calculated between the areas of prediction and ground truth lanes drawn at a given width.
Unilateral Chamfer Distance (CD) between the pairs above the IoU threshold is then calculated as the matching error, and a true positive is recognized if its CD error is below the specified threshold $\tau_{CD}$.
F1 score, precision, recall, and CD error are reported on ONCE-3DLanes as evaluation metrics.

\subsubsection{Implementation Details}
We adopt ResNet-18~\cite{resnet} and ResNet-50~\cite{resnet} as the backbones.
Due to the requirement for feature resolution for the lane detection task, the downsampling stride of the last two stages of our backbone is changed to $1$ and the $3\times 3$ convolutions are also replaced with dilated convolutions to maintain the receptive field.
The meta prototypes are uniformly initialized and their number $M_x$, $M_\phi$, and $M_\theta$ are set to $30$, $15$, and $5$ respectively.
The total number of sparse anchors $N_a$ is set to $30$.
The number of iterative stages is set to $4$ in our final implementation, where the 1st and 2nd stages are both conducted on $\mathbf{F}^5$.
According to their y-coordinate ranges, the number of sampled anchor points $N$ is set to $20$ on ApolloSim and OpenLane datasets, and $10$ for ONCE-3DLanes.
Two different input image resolutions are adopted in our paper to explore, including $360\times 480$ and $720\times 960$.
For the LiDAR modality, we adopt SECOND~\cite{yan2018second} as the voxel-based point encoder and PointPillars~\cite{lang2019pointpillars} as the pillar-based point encoder.
The raw point clouds are divided into regular voxels/pillars before being fed into point encoder.
For SECOND as the point encoder, the range of point cloud is clipped into [$-30.4m$, $30.4m$] for X-axis, [$-1.6m$, $75.2m$] for Y-axis, and [$-4m$, $4m$] for Z-axis.
The input voxel size is set to ($0.32m$, $0.32m$, $0.15m$).
For PointPillars as the point encoder, the range of point cloud is clipped into [$-30m$, $30m$] for X-axis, [$-0.64m$, $74.88m$] for Y-axis, and [$-2m$, $4m$] for Z-axis, and the grid size is set to ($0.32m$, $0.32m$).

During training, $\tau$ in EW loss is set to $0.1$ based on the general road construct standards and experimental results.
The coefficients in matching cost are set to $1$ and $3$ for $\beta_{cls}$ and $\beta_{dis}$, and the loss coefficients are set to $1$, $1$, and $0.1$ for $\lambda_{cls}$, $\lambda_{reg}$, $\lambda_{EW}$ respectively.
Adam optimizer with weight decay set as $1e^{-4}$ and initial learning rate as $1e^{-4}$ is used.
The batch size is $16$ and the model is trained for $50K$ iterations on ApolloSim with $1$ NVIDIA A100 GPU.
For OpenLane and ONCE-3DLanes datasets, the batch size is $64$ and the model is trained for $60k$ iterations with $4$ NVIDIA A100 GPUs.

\subsection{Comparison with State-of-the-Art Methods}

\begin{table*}[!htbp]
\begin{center}
\caption{Comparison with state-of-the-art methods on the OpenLane validation set. ``*'' indicates results on the original version of the OpenLane dataset. ``R18'' and ``R50'' are short for ResNet-18 and ResNet-50 image backbones. ``PP'' and ``SE'' are short for PointPillars and SECOND LiDAR backbones. ``$\dagger$'' indicates larger image resolution (i.e., $720\times 960$). ``C'' and ``L'' denotes camera and LiDAR modalities. ``CAcc'' means category accuracy. ``E$_x$'' and ``E$_z$'' are short for x and z errors respectively, and ``N'' and ``F'' for near and far respectively. ``TP'' denotes the throughput.}
\resizebox{\linewidth}{!}{
\begin{tabular}{c|c|cccccc|cc}
\toprule
\textbf{Method}  & \textbf{Modality} & \textbf{F1(\%)$\uparrow$} & \textbf{CAcc(\%)$\uparrow$} & \textbf{E$_x$/N(m) $\downarrow$} & \textbf{E$_x$/F(m) $\downarrow$} & \textbf{E$_z$/N(m) $\downarrow$} & \textbf{E$_z$/F(m) $\downarrow$} & \textbf{FPS $\uparrow$} & \textbf{TP $\uparrow$} \\ 
\hline
3D-LaneNet*\cite{3dlanenet}~\tiny{CVPR2019}  & C & 44.1 & - & 0.479 & 0.572 & 0.367 & 0.443 & 67.5 & 204.4 \\
GenLaneNet*\cite{genlanenet}~\tiny{ECCV2020} & C & 32.3 & - & 0.591 & 0.684 & 0.411 & 0.521 & 16.6 & 24.3 \\
PersFormer*\cite{persformer}~\tiny{ECCV2022} & C & 50.5 & 92.3 & 0.485 & 0.553 & 0.364 & 0.431 & 18.1 & 26.7 \\
CurveFormer*\cite{curveformer}~\tiny{ICRA2023} & C & 50.5 & - & 0.340 & 0.772 & 0.207 & 0.651 & - & - \\
MapTRv2~\cite{MapTR,maptrv2}~\scriptsize{(R50-GKT)}~\tiny{ICLR2023} & C & 53.0 &  88.0 & 0.288 & 0.321 & 0.077 & 0.109 & 26.1 & 100.8 \\
MapTRv2~\cite{MapTR,maptrv2}~\scriptsize{(R50-BEVFormer)}~\tiny{ICLR2023} & C & 53.6 & 88.9 & 0.267 & 0.312 & 0.074 & 0.105 & 26.0 & 94.8 \\
\hline
Anchor3DLane~\scriptsize{(R18)}*\cite{anchor3dlane}~\tiny{CVPR2023} & C & 53.7 & 90.9 & 0.276 & 0.311 & 0.107 & 0.138 & \textbf{72.1} & \textbf{401.0} \\
Anchor3DLane~\scriptsize{(R18)}~\cite{anchor3dlane}~\tiny{CVPR2023} & C & 53.6 & 89.2 & 0.279 & 0.301 & 0.085 & 0.117 & \textbf{72.1} & \textbf{401.0} \\
Anchor3DLane~\scriptsize{(R50)}$\dagger$~\cite{anchor3dlane}~\tiny{CVPR2023} & C & 57.5 & 91.6 & 0.233 & 0.246 & 0.080 & 0.106 & 32.7 & 57.8 \\
\rowcolor{mygray}
Anchor3DLane++~\scriptsize{(R18)} & C & 57.9 & 91.4 & 0.232 & 0.265 & 0.076 & 0.102 & 38.1 & 282.4 \\
\rowcolor{mygray}
Anchor3DLane++~\scriptsize{(R50)} & C & 59.4 & 92.6 & 0.227 & 0.244 & 0.075 & \textbf{0.100} & 30.5 & 149.5 \\
\rowcolor{mygray}
Anchor3DLane++~\scriptsize{(R50)}$\dagger$ & C & \textbf{62.4} & \textbf{93.4} & \textbf{0.202} & \textbf{0.237} & \textbf{0.073} & \textbf{0.100} & 22.9 & 49.4 \\
\hline
MapTRv2~\cite{MapTR,maptrv2}~\scriptsize{(R50-GKT)}~\tiny{ICLR2023} & C+L & 54.8 & 89.5 & 0.217 & 0.251 & 0.037 & 0.061 & 8.0 & 13.8 \\
MapTRv2~\cite{MapTR,maptrv2}~\scriptsize{(R50-BEVFormer)}~\tiny{ICLR2023} & C+L & 55.2 & 88.8 & 0.221 & 0.236 & 0.037 & 0.073 & 7.9 & 13.5 \\
M$^2$-3DLaneNet*~\cite{m2net}~\tiny{arXiv2022} & C+L & 55.5 & - & 0.431 & 0.487 & 0.327 & 0.401 & - & - \\
\hline
\rowcolor{mygray}
Anchor3DLane++~\scriptsize{(R18+SE)}* & C+L & 60.4 & 92.6 & 0.198 & 0.201 & 0.065 & 0.084 & 15.9 & 53.2 \\ 
\rowcolor{mygray}
Anchor3DLane++~\scriptsize{(R18+SE)} & C+L & 59.8 & 92.6 & 0.167 & 0.170 & 0.035 & 0.060 & 15.9 & 53.2 \\
\rowcolor{mygray}
Anchor3DLane++~\scriptsize{(R50+SE)} & C+L & 61.4 & 92.9 & 0.149 & 0.160 & \textbf{0.033} & 0.058 & 13.8 & 45.4 \\
\rowcolor{mygray}
Anchor3DLane++~\scriptsize{(R50+SE)}$\dagger$ & C+L & \textbf{62.9} & \textbf{93.6} & \textbf{0.134} & \textbf{0.137} & \textbf{0.033} & \textbf{0.057} & 12.4 & 27.2 \\
\rowcolor{mygray}
Anchor3DLane++~\scriptsize{(R18+PP)} & C+L & 60.0 & 92.5 & 0.164 & 0.177 & 0.049 & 0.082 & \textbf{21.6} & \textbf{85.4} \\
\rowcolor{mygray}
Anchor3DLane++~\scriptsize{(R50+PP)} & C+L & 61.1 & 93.1 & 0.147 & 0.165 & 0.055 & 0.091 & 17.8 & 79.0 \\
\rowcolor{mygray}
Anchor3DLane++~\scriptsize{(R50+PP)$\dagger$} & C+L & \textbf{62.9} & \textbf{93.6} & 0.148 & 0.152 & 0.047 & 0.079 & 15.7 & 36.8 \\
\bottomrule
\end{tabular}}
\label{tab:sota-openlane}
\end{center}
\end{table*}

\begin{table*}[!htbp]
\begin{center}
\caption{Comparison with state-of-the-art methods on OpenLane validation set. F1 score is presented for each scenario. ``*'' indicates results on the original version of the OpenLane dataset. ``R18'' and ``R50'' are short for ResNet-18 and ResNet-50 image backbones. ``PP'' and ``SE'' are short for PointPillars and SECOND LiDAR backbones. ``$\dagger$'' indicates larger image resolution (i.e., $720\times 960$). ``C'' and ``L'' denotes camera and LiDAR modalities.}
\resizebox{\linewidth}{!}{
\begin{tabular}{c|c|ccccccc}
\toprule
\textbf{Method} & \textbf{Modality} & \textbf{All} & \textbf{Up \& Down} & \textbf{Curve} & \textbf{Extreme Weather} & \textbf{Night} & \textbf{Intersection} & \textbf{Merge \& Split} \\ 
\hline
3D-LaneNet*~\cite{3dlanenet}~\tiny{CVPR2019} & C & 44.1 & 40.8 & 46.5 & 47.5 & 41.5 & 32.1 & 41.7 \\
GenLaneNet*~\cite{genlanenet}~\tiny{ECCV2020} & C & 32.3 & 25.4 & 33.5 & 28.1 & 18.7 & 21.4 & 31.0 \\
PersFormer*~\cite{persformer}~\tiny{ECCV2022} & C & 50.5 & 42.4 & 55.6 & 48.6 & 46.6 & 40.0 & 50.7 \\
CurveFormer*~\cite{curveformer}~\tiny{ICRA2023} & C & 50.5 & 45.2 & 56.6 & 49.7 & 49.1 & 42.9 & 45.4 \\
MapTRv2~\cite{MapTR,maptrv2}~\scriptsize{(R50-GKT)}~\tiny{ICLR2023} & C & 53.0 & 48.6 & 53.8 & 51.8 & 48.3 & 42.5 & 53.3 \\
MapTRv2~\cite{MapTR,maptrv2}~\scriptsize{(R50-BEVFormer)}~\tiny{ICLR2023} & C & 53.6 & 50.0 & 53.2 & 54.9 & 51.3 & 43.1 & 53.1 \\
\hline
Anchor3DLane~\scriptsize{(R18)}*~\cite{anchor3dlane}~\tiny{CVPR2023} & C & 53.7 & 46.7 & 57.2 & 52.5 & 47.8 & 45.4 & 51.2 \\
Anchor3DLane~\scriptsize{(R18)}~\cite{anchor3dlane}~\tiny{CVPR2023} & C & 53.6 & 47.8 & 58.1 & 50.9 & 49.0 & 45.8 & 51.5 \\
Anchor3DLane~\scriptsize{(R50)}$\dagger$~\cite{anchor3dlane}~\tiny{CVPR2023} & C & 57.5 & 52.7 & 60.8 & 56.2 & 54.7 & 49.8 & 56.0 \\
\rowcolor{mygray}
Anchor3DLane++~\scriptsize{(R18)} & C & 57.9 & 48.4 & 64.0 & 54.8 & 52.6 & 48.5 & 56.1 \\
\rowcolor{mygray}
Anchor3DLane++~\scriptsize{(R50)} & C & 59.4 & 49.9 & 66.1 & 55.5 & 52.5 & 50.3 & 57.4 \\
\rowcolor{mygray}
Anchor3DLane++~\scriptsize{(R50)}$\dagger$ & C & \textbf{62.4} & \textbf{54.1} & \textbf{68.4} & \textbf{58.3} & \textbf{55.4} & \textbf{53.1} & \textbf{61.1} \\
\hline
MapTRv2~\cite{MapTR,maptrv2}~\scriptsize{(R50-GKT)}~\tiny{ICLR2023} & C+L & 54.8 & 50.3 & 54.7 & 54.8 & 53.8 & 45.1 & 53.5 \\
MapTRv2~\cite{MapTR,maptrv2}~\scriptsize{(R50-BEVFormer)}~\tiny{ICLR2023} & C+L & 55.2 & 51.6 & 55.1 & 52.7 & 52.7 & 45.2 & 54.3 \\
M$^2$-3DLaneNet*~\cite{m2net}~\tiny{arXiv2022} & C+L & 55.5 & 53.4 & 60.7 & 56.2 & 51.6 & 43.8 & 51.4 \\
\hline
\rowcolor{mygray}
Anchor3DLane++~\scriptsize{(R18+SE)}* & C+L & 60.4 & 53.8 & 66.3 & 60.2 & 55.9 & 50.3 & 58.1 \\
\rowcolor{mygray}
Anchor3DLane++~\scriptsize{(R18+SE)} & C+L & 59.8 & 54.2 & 67.6 & 55.8 & 55.0 & 50.6 & 57.5 \\
\rowcolor{mygray}
Anchor3DLane++~\scriptsize{(R50+SE)} & C+L & 61.4 & 55.5 & 68.7 & 55.0 & 55.1 & 51.7 & 60.8 \\
\rowcolor{mygray}
Anchor3DLane++~\scriptsize{(R50+SE)}$\dagger$ & C+L & \textbf{62.9} & \textbf{56.6} & \textbf{69.9} & 59.7 & 57.1 & \textbf{54.0} & \textbf{61.5} \\
\rowcolor{mygray}
Anchor3DLane++~\scriptsize{(R18+PP)} & C+L & 60.0 & 53.4 & 67.0 & 59.5 & 56.4 & 50.4 & 57.9 \\
\rowcolor{mygray}
Anchor3DLane++~\scriptsize{(R50+PP)} & C+L & 61.1 & 54.9 & 68.0 & 57.6 & \textbf{57.7} & 51.5 & 59.6 \\
\rowcolor{mygray}
Anchor3DLane++~\scriptsize{(R50+PP)$\dagger$} & C+L & \textbf{62.9} & 56.5 & 69.2 & \textbf{60.5} & \textbf{57.7} & 53.5 & 60.2 \\
\bottomrule
\end{tabular}}
\label{tab:scene_openlane}
\end{center}
\end{table*}

\noindent \textbf{Results on OpenLane}.
Experimental results on OpenLane dataset\footnote{Since the annotation of OpenLane dataset has been refined since 2022/11, all experiments in this paper are based on the refined data version, and the quantitative results of our conference version are also updated accordingly.} are presented in Table~\ref{tab:sota-openlane} and Table~\ref{tab:scene_openlane}.
Our original Anchor3DLane already outperforms the previous state-of-the-art method, MapTRv2\footnote{We utilize the official code of MapTRv2 and adapt it to 3D lane detection. All hyperparameters are kept the same as the default settings, except that 3D lane NMS is incorporated for post-processing.}, by $3.9\%$ F1 score improvement, with ResNet50 as the backbone, which significantly demonstrates the advantages of direct regression from FV representations.
It is also worth mentioning that our extended method Anchor3DLane++ has achieved substantial performance improvements across all metrics compared to the previous conference version.
Specifically, benefiting from the sample-adaptive anchor generation process and cross-layer iterative refinement, Anchor3DLane++ significantly outperforms Anchor3DLane in Curve ($+5.9\%$ F1 score) and Merge\&Split scenarios ($+4.6\%$ F1 score).
To verify the adaptability and performance potential of our method, we conduct experiments using ResNet-50 as the backbone with a larger input resolution for both Anchor3DLane and Anchor3DLane++, which further boosts the overall performance significantly.
Due to the fine-grained nature of the lane detection task, doubling the input resolution results in more performance improvement than using a larger image backbone.

Apart from the camera-only results, we also show the experimental results of camera-LiDAR fusion settings in Table~\ref{tab:sota-openlane} and Table~\ref{tab:scene_openlane}.
Compared with the camera-only settings, incorporating the LiDAR modality into Anchor3DLane++ significantly reduces the x and z errors and improves the F1 score.
Additionally, camera-LiDAR fusion brings notable performance gains in the Up\&Down scenarios, further demonstrating the advantage of the LiDAR modality in 3D position perception. 
However, due to the high results from large-resolution image inputs, the performance gains from camera-LiDAR fusion are not as significant as in the small-resolution experiments as shown in the last rows of Table~\ref{tab:sota-openlane} and Table~\ref{tab:scene_openlane}.
We also compare Anchor3DLane++ with other camera-LiDAR fusion methods, including MapTRv2 and M$^2$-3DLaneNet. 
Our lightest model (R18) has already surpassed M$^2$-3DLaneNet across all metrics by large margins, which further demonstrates the intrinsic advantages of Anchor3DLane++.

\begin{table*}[t]
\begin{center}
\caption{Comparison with state-of-the-art methods on the ApolloSim dataset with three different split settings. ``E$_x$'' and ``E$_z$'' are short for x and z errors respectively, and ``N'' and ``F'' for near and far respectively. ``R18'' and ``R50'' are short for ResNet-18 and ResNet-50 backbones. ``$\dagger$'' indicates larger image resolution (i.e., $720\times 960$).}
\resizebox{1\linewidth}{!}{
\begin{tabular}{c|c|cccccc}
\toprule
\textbf{Scene} & \textbf{Method} & \textbf{AP(\%)$\uparrow$} & \textbf{F1(\%)$\uparrow$} & \textbf{E$_x$/N(m) $\downarrow$} & \textbf{E$_x$/F(m)} $\downarrow$ & \textbf{E$_z$/N(m) $\downarrow$} & \textbf{E$_z$/F(m) $\downarrow$} \\
\hline
\multirow{7}{*}{Balanced Scene} & 3DLaneNet~\cite{3dlanenet}~\tiny{CVPR2019} & 89.3 & 86.4 & 0.068 & 0.477 & 0.015 & \textbf{0.202} \\
& Gen-LaneNet~\cite{genlanenet}~\tiny{ECCV2020} & 90.1 & 88.1 & 0.061 & 0.496 & 0.012 & 0.214 \\
& CLGo~\cite{clgo}~\tiny{AAAI2022} & 94.2 & 91.9 & 0.061 & 0.361 & 0.029 & 0.250 \\
& PersFormer~\cite{persformer}~\tiny{ECCV2022} & - & 92.9 & 0.054 & 0.356 & 0.010 & 0.234 \\
& WS-3D-Lane~\cite{ai2023ws}~\tiny{ICRA2023} & 95.7 & 93.5 & 0.027 & 0.321 & 0.006 & 0.215 \\
& CurveFormer~\cite{curveformer}~\tiny{ICRA2023} & 97.3 & 95.8 & 0.079 & 0.326 & 0.018 & 0.219 \\
\cline{2-8}
& Anchor3DLane~\cite{anchor3dlane}~\tiny{CVPR2023} & 97.1 & 95.4 & 0.045 & 0.300 & 0.016 & 0.223 \\
& \cellcolor{mygray}Anchor3DLane++~\scriptsize{(R18)} & \cellcolor{mygray}\textbf{97.6} & \cellcolor{mygray}96.3 & \cellcolor{mygray}0.027 & \cellcolor{mygray}0.268 & \cellcolor{mygray}0.011 & \cellcolor{mygray}0.215 \\
& \cellcolor{mygray}Anchor3DLane++~\scriptsize{(R50)}$\dagger$ & \cellcolor{mygray}\textbf{97.6} & \cellcolor{mygray}\textbf{96.5} & \cellcolor{mygray}\textbf{0.022}& \cellcolor{mygray}\textbf{0.234}& \cellcolor{mygray}\textbf{0.009} & \cellcolor{mygray}\textbf{0.204} \\
\hline
\multirow{7}{*}{Rare Subset} & 3DLaneNet~\cite{3dlanenet}~\tiny{CVPR2019} & 74.6 & 72.0 & 0.166 & 0.855 & 0.039 & \textbf{0.521} \\
& Gen-LaneNet~\cite{genlanenet}~\tiny{ECCV2020} & 79.0 & 78.0 & 0.139 & 0.903 & 0.030 & 0.539 \\
& CLGo~\cite{clgo}~\tiny{AAAI2022} & 88.3 & 86.1 & 0.147 & 0.735 & 0.071 & 0.609 \\
& PersFormer~\cite{persformer}~\tiny{ECCV2022} & - & 87.5 & 0.107 & 0.782 & 0.024 & 0.602 \\
& CurveFormer~\cite{curveformer}~\tiny{ICRA2023} & 97.1 & 95.6 & 0.182 & 0.737 & 0.039 & 0.561 \\
\cline{2-8}
& Anchor3DLane~\cite{anchor3dlane}~\tiny{CVPR2023}  & 95.9 & 94.4 & 0.082 & 0.699 & 0.030 & 0.580 \\
& \cellcolor{mygray}Anchor3DLane++~\scriptsize{(R18)} & \cellcolor{mygray}\textbf{97.7} & \cellcolor{mygray}\textbf{96.4} & \cellcolor{mygray}0.050 & \cellcolor{mygray}0.617 & \cellcolor{mygray}0.019 & \cellcolor{mygray}0.551 \\
& \cellcolor{mygray}Anchor3DLane++~\scriptsize{(R50)}$\dagger$ & \cellcolor{mygray}97.6 & \cellcolor{mygray}\textbf{96.4} & \cellcolor{mygray}\textbf{0.043}& \cellcolor{mygray}\textbf{0.580}& \cellcolor{mygray}\textbf{0.017} & \cellcolor{mygray}\textbf{0.529} \\
\hline
\multirow{7}{*}{Visual Variations} & 3D-LaneNet~\cite{3dlanenet}~\tiny{CVPR2019} & 74.9 & 72.5 & 0.115 & 0.601 & 0.032 & 0.230 \\
& Gen-LaneNet~\cite{genlanenet}~\tiny{ECCV2020} & 87.2 & 85.3 & 0.074 & 0.538 & 0.015 & 0.232 \\
& CLGo~\cite{clgo}~\tiny{AAAI2022} & 89.2 & 87.3 & 0.084 & 0.464 & 0.045 & 0.312 \\
& PersFormer~\cite{persformer}~\tiny{ECCV2022} & - & 89.6 & 0.074 & 0.430 & 0.015 & 0.266 \\
& CurveFormer~\cite{curveformer}~\tiny{ICRA2023} & 93.0 & 90.8 & 0.125 & 0.410 & 0.028 & 0.254 \\
\cline{2-8} 
& Anchor3DLane~\cite{anchor3dlane}~\tiny{CVPR2022} & 92.5 & 91.8 & 0.047 & 0.327 & 0.019 & 0.219 \\
& \cellcolor{mygray}Anchor3DLane++~\scriptsize{(R18)} & \cellcolor{mygray}95.1 & \cellcolor{mygray}92.7 & \cellcolor{mygray}0.045 & \cellcolor{mygray}0.371 & \cellcolor{mygray}0.019 & \cellcolor{mygray}0.250 \\
& \cellcolor{mygray}Anchor3DLane++~\scriptsize{(R50)}$\dagger$ & \cellcolor{mygray}\textbf{97.1} & \cellcolor{mygray}\textbf{95.3} & \cellcolor{mygray}\textbf{0.035} & \cellcolor{mygray}\textbf{0.292} & \cellcolor{mygray}\textbf{0.012} & \cellcolor{mygray}\textbf{0.229} \\
\bottomrule     
\end{tabular}}
\label{tab:sota-apollo}
\end{center}
\end{table*}

\noindent \textbf{Results on ApolloSim}.
We present experimental results under three different split settings (\textit{i.e.}, balanced scene, rare subset, and visual variations) of the ApolloSim dataset in Table~\ref{tab:sota-apollo}.
With the simple design, our original Anchor3DLane achieves significantly higher AP and F1 scores on all three splits than previous methods.
Compared with the concurrent work~\cite{curveformer}, Anchor3DLane still achieves comparable AP and F1 scores but much lower x errors, indicating the superiority of explicit 3D lane modeling.
Anchor3DLane++ achieves further performance improvements over the previous conference version, especially in terms of F1 score and AP.
Furthermore, by utilizing a larger backbone and input resolution, we observe an obvious reduction in x and z errors, demonstrating the capability of our Anchor3DLane++.

\noindent \textbf{Results on ONCE-3DLanes}.
Experimental results on the ONCE-3DLanes dataset are illustrated in Table~\ref{tab:once}.
Given that the ONCE-3DLanes dataset does not provide camera extrinsic, 3D anchors are defined in the camera coordinate system, and predictions are also made in the same space.
Our original Anchor3DLane already outperforms previous methods, including SALAD and PersFormer, which indicates that our 3D anchors can adapt to different 3D coordinate systems.
Furthermore, the Anchor3DLane++ achieves much higher F1 scores than the previous state-of-the-art method~\cite{ai2023ws} with a lightweight image backbone.

\begin{figure}[!htbp]
    \centering
    \includegraphics[width=0.9\linewidth]{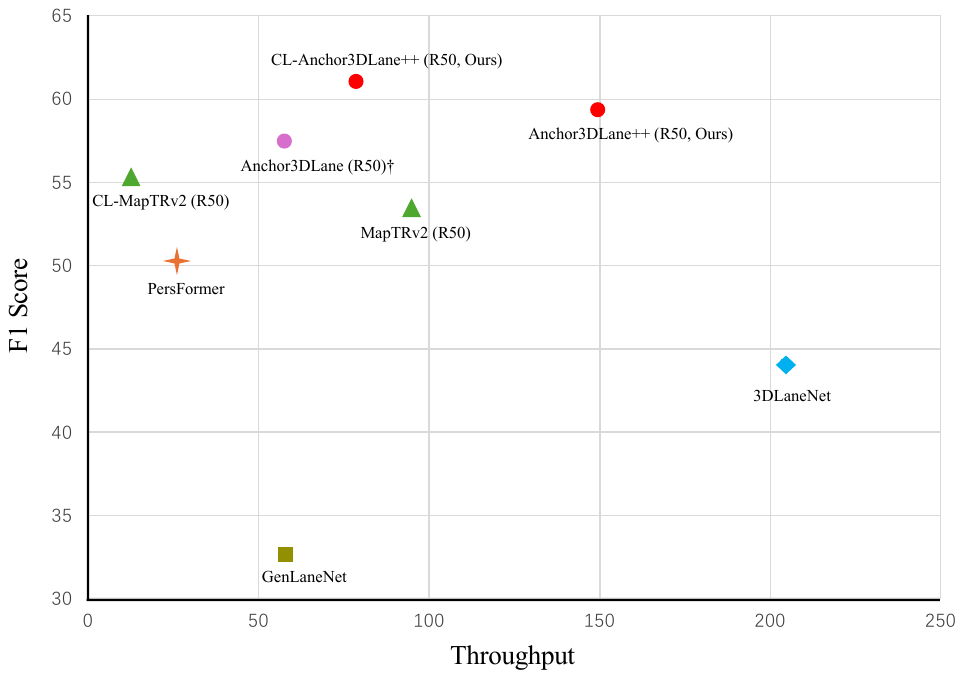}
    \caption{Comparison of F1 score vs. throughput for different methods.}
    \label{fig:throughput}
\end{figure}

\noindent \textbf{Computational Efficiency}.
In the last two columns of Table~\ref{tab:sota-openlane}, we test the frames per second (FPS) and throughput (TP) of different methods on a single NVIDIA A100 GPU, with FPS tested with a batch size of $1$ and averaged over $2000$ repetitions, and TP tested with a batch size of $32$ and averaged over $200$ repetitions.
We also visualize the F1 scores and TP of different methods in Fig.~\ref{fig:throughput} for a more straightforward comparison.
Both Anchor3DLane (our conference version) and our camera-only Anchor3DLane++ achieve high TP (\textit{e.g.}, $401.0$ and $282.4$, respectively), meeting the requirements for real-time applications.
It is shown that compared with MapTRv2, our Anchor3DLane++ achieves a higher performance ($59.4$ vs. $53.6$ F1 score) with much faster speed ($149.5$ vs. $94.8$ TP), demonstrating the computational efficiency of our method.
Although incorporating the LiDAR modality into Anchor3DLane++ decreases the speed due to the additional computational cost of LiDAR point processing, it remains much faster than MapTRv2 (\textit{e.g.}, $79.0$ vs. $13.8$ TP).

\begin{table}[!htbp]
\begin{center}
\caption{Comparison with state-of-the-art methods on the ONCE-3DLanes validation set. Results under $\tau_{CD}=0.3$ are displayed here. ``P'', ``R'', and ``CDE'' are short for precision, recall, and CD error respectively. ``R18'' and ``R50'' are short for ResNet-18 and ResNet-50 backbones. ``$\dagger$'' indicates larger image resolution (i.e., $720\times 960$).}
\resizebox{\linewidth}{!}{
\begin{tabular}{c|cccc}
\toprule
\textbf{Method} & \textbf{F1(\%)$\uparrow$} & \textbf{P(\%)$\uparrow$ }& \textbf{R(\%)$\uparrow$} & \textbf{CDE(m)$\downarrow$} \\ 
\hline
3D-LaneNet~\cite{3dlanenet}~\tiny{CVPR2019} & 44.73 & 61.46 & 35.16 & 0.127 \\
Gen-LaneNet~\cite{genlanenet}~\tiny{ECCV2020} & 45.59 & 63.95 & 35.42 & 0.121\\
SALAD~\cite{once}~\tiny{CVPR2022} & 64.07 & 75.90 & 55.42 & 0.098 \\
PersFormer~\cite{persformer}~\tiny{ECCV2022} & 74.33 & 80.30 & 69.18 & 0.074 \\
WS-3D-Lane~\cite{ai2023ws}~\tiny{ICRA2023} & 77.02 & \textbf{84.51} & 70.75 & 0.058 \\
\hline
Anchor3DLane~\cite{anchor3dlane}~\tiny{CVPR2023} & 74.87 & 80.85 & 69.71 & 0.060 \\
\rowcolor{mygray}
Anchor3DLane++~\scriptsize{(R18)} & 79.55 & 82.67 & 76.67 & 0.059 \\
\rowcolor{mygray}
Anchor3DLane++~\scriptsize{(R50)}$\dagger$ & \textbf{81.25} & 84.18 & \textbf{78.52} & \textbf{0.055} \\
\bottomrule
\end{tabular}}
\label{tab:once}
\end{center}
\end{table}

\subsection{Ablation Study}
To evaluate different designs of our Anchor3DLane++, we conduct ablation studies on the OpenLane dataset with ResNet-18 as the backbone.
To illustrate the impact of each design on performance more intuitively, most of our ablation experiments are based on a single stage of iterative regression apart from ablations on the number of iterative stages.
\begin{table*}[!htbp]
\begin{center}
\caption{Ablation studies on each component of Anchor3DLane++. ``L'' and ``G'' denote equal-width regularization between adjacent 3D proposals and all pairs of proposals respectively.}
\resizebox{\linewidth}{!}{
\begin{tabular}{ccc|cc|cccccc}
\toprule
\textbf{Sparse} & \textbf{Dynamic} & \textbf{PAAG} & \textbf{EW Loss (L)} & \textbf{EW Loss (G)} & \textbf{F1(\%)} & \textbf{CAcc} & \textbf{E$_x$/N(m)}& \textbf{E$_x$/F(m)} & \textbf{E$_z$/N(m)} & \textbf{E$_z$/F(m)} \\
\hline
\checkmark & & & & & 52.3 & 87.3 & 0.366 & 0.369 & 0.085 & 0.119 \\
\checkmark & \checkmark & & & &  53.0 & 89.0 & 0.335 & 0.340 & 0.083 & 0.113 \\
\checkmark & \checkmark & \checkmark & & & 54.3 & \textbf{89.9} & 0.295 & 0.305 & 0.079 & 0.112 \\
\checkmark & \checkmark & \checkmark & \checkmark & & 54.5 & 89.6 & 0.291 & 0.297 & 0.081 & 0.112 \\
\checkmark & \checkmark & \checkmark &  & \checkmark & \textbf{54.9} & \textbf{89.9} & \textbf{0.289} & \textbf{0.296} & \textbf{0.080} & \textbf{0.110} \\
\bottomrule     
\end{tabular}}
\label{tab:ab}
\end{center}
\end{table*}

\begin{table*}[!htbp]
\begin{center}
\caption{Ablation study on camera-LiDAR fusion. ``Eval Range'' indicates the range along the Y-axis over which the evaluation metrics are calculated.}
\resizebox{.9\linewidth}{!}{
\begin{tabular}{cc|c|cccccc}
\toprule
\textbf{Camera}  & \textbf{LiDAR} & \textbf{Eval Range} & \textbf{F1(\%)} & \textbf{CAcc} & \textbf{E$_x$/N(m)}& \textbf{E$_x$/F(m)} & \textbf{E$_z$/N(m)} & \textbf{E$_z$/F(m)} \\
\hline
\checkmark & & \multirow{3}{*}{0 $-$ 75m} & 58.2 & 90.4 & 0.269 & 0.282 & 0.079 & 0.110 \\
 & \checkmark &  & 56.0 & 85.3 & 0.244 & 0.213 & \textbf{0.039} & \textbf{0.060} \\
\checkmark & \checkmark &  & \textbf{60.9} & \textbf{91.8} & \textbf{0.198} & \textbf{0.192} & 0.042 & 0.066 \\
\hline
\checkmark & & \multirow{2}{*}{0 $-$ 100m} & 54.9 & 89.9 & 0.289 & 0.296 & 0.080 & 0.110  \\
\checkmark & \checkmark  &  & \textbf{57.4} & \textbf{91.5} & \textbf{0.212} & \textbf{0.207} & \textbf{0.042} & \textbf{0.068} \\
\bottomrule     
\end{tabular}}
\label{tab:LiDAR}
\end{center}
\end{table*}

\begin{table}[!htbp]
\begin{center}
\caption{Comparison between sampling anchor features from FV features and BEV features, where BEV features are extracted by different BEV encoders.}
\resizebox{\linewidth}{!}{
\begin{tabular}{c|ccccc}
\toprule
\textbf{Feature Type} & \textbf{F1(\%)} & \textbf{E$_x$/N(m)}& \textbf{E$_x$/F(m)} & \textbf{E$_z$/N(m)} & \textbf{E$_z$/F(m)} \\
\hline
IPM Image~\cite{mallot1991inverse} & 53.4 & 0.321 & 0.316 & 0.083 & 0.140 \\
\hline
IPM Feature~\cite{mallot1991inverse} & 53.5 & 0.312 & 0.314 & 0.081 & 0.127 \\
LSS~\cite{philion2020lift,huang2022bevpoolv2} & 53.5 & 0.333 & 0.336 & 0.088 & 0.125 \\
GKT~\cite{chen2022efficient} & 53.7 & 0.324 & 0.317 & 0.086 & 0.126 \\
BEVFormer~\cite{li2022bevformer,zhu2020deformable} & 54.2 & 0.308 & 0.321 & 0.082 & 0.113 \\
\hline
FV Feature & \textbf{54.9} & \textbf{0.289} & \textbf{0.296} & \textbf{0.080} & \textbf{0.110} \\
\bottomrule     
\end{tabular}}
\label{tab:bev}
\end{center}
\end{table}

\noindent \textbf{Sampling anchor features from FV features.} 
We first compare the experimental results of sampling anchor features from FV features and BEV features respectively to verify the advantages of the former.
The results are shown in Table~\ref{tab:bev}.
Different ways to acquire BEV features are investigated, including taking the warped BEV image as input of the image encoder (row 1), warping FV features into BEV features using IPM~\cite{mallot1991inverse}(row 2), and extracting BEV features through other advanced BEV encoders such as BEVFormer~\cite{li2022bevformer}, LSS~\cite{philion2020lift} and GKT~\cite{chen2022efficient} (row3-5).
All the other architectures and training settings are kept the same as our single-stage Anchor3DLane++ (row 6).
It is shown that due to its layer-by-layer feature refinement mechanism, BEVFormer provides the best BEV feature representations for anchor feature sampling and achieves the highest overall performance over other BEV encoders.
However, sampling anchor features from FV features still yields the best performances, proving that the context information maintained in the raw FV features facilitates the prediction of 3D lanes.

\noindent \textbf{Prototype-based adaptive anchor generation.}
We explore different ways of sparse anchor generation to verify the effectiveness of PAAG.
As shown in Table~\ref{tab:ab}, due to the severely insufficient coverage of sparse anchors across different road conditions, naively sparsifying 3D anchors leads to a noticeable decline in performance compared with the dense counterpart in Table~\ref{tab:sota-openlane}, especially in x errors.
By setting these sparse anchors as learnable parameters and allowing them to be dynamically updated during training, performance can be somewhat improved, while the x errors remain high.
In the third row, incorporating our PAAG module to generate sample-adaptive sparse anchors using anchor meta prototypes yields significant improvement in both F1 scores and x/z errors, which already surpasses the previous Anchor3DLane.
From these results, it is shown that our PAAG can generate sparse anchors aligned with the input image, thus mitigating the insufficient coverage caused by anchor sparsification.

\noindent \textbf{Equal-Width loss.}
We further verify the effectiveness of our Equal-Width loss in Table~\ref{tab:ab}.
In addition to constraining the width consistency among all proposal pairs globally, \textit{i.e.}, EW Loss (G), we also explore the local optimization by constraining only the adjacent proposals, \textit{i.e.}, EW Loss (L), by setting the proposal index $j'$ to $j+1$ in Equation~\ref{eq:ewl}.
It is shown that local optimization is too weak and has limited contribution to performance improvement.
By applying the equal-width regularization among all the proposal pairs, a significant improvement in the F1 score is observed, indicating that the equal-width property does benefit 3D lane prediction.

\begin{figure*}[!htbp]
    \centering
    \includegraphics[width=0.95\linewidth]{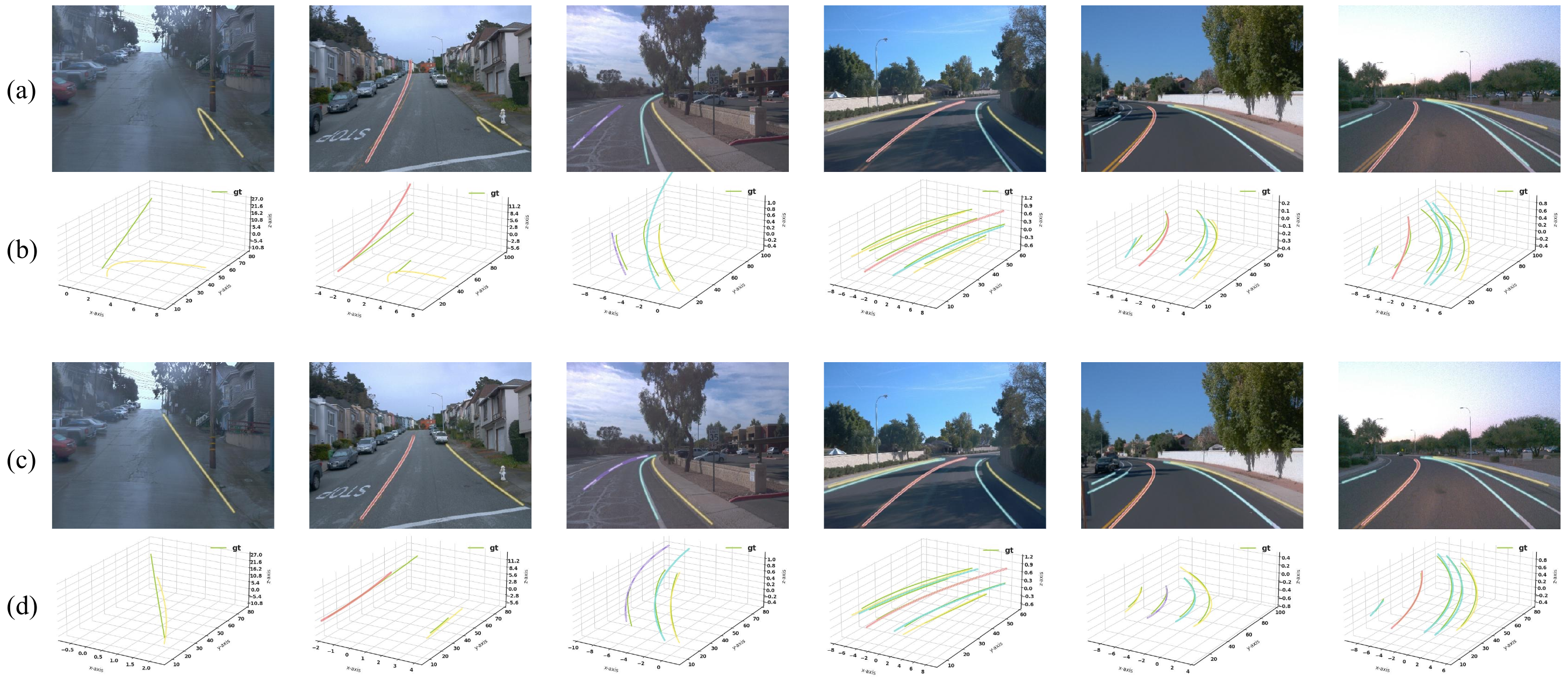}
    \caption{Qualitative comparison between PersFormer and our Anchor3DLane++ on the OpenLane dataset. (a) Projected predictions of PersFormer on 2D images. (b) 3D predictions of PersFormer. (c) Projected predictions of Anchor3DLane++ on 2D images. (d) 3D predictions of Anchor3DLane++.}
    \label{fig:vis_lane}
\end{figure*} 
\noindent \textbf{Camera-LiDAR fusion.}
We adopt SECOND as the point encoder and illustrate the experimental results of different modalities in Table~\ref{tab:LiDAR}.
Since the LiDAR points extend up to $75.2m$ along the y-axis, we calculate the evaluation metrics within the range of $0-75m$ when comparing with the LiDAR-only setting.
It is observed from the upper part of Table~\ref{tab:LiDAR} that due to the precise depth perception ability of the LiDAR modality, the LiDAR-only setting produces lower x/z errors. 
However, because it is weaker in semantic perception compared with the camera modality, LiDAR-only significantly underperforms camera-only in terms of classification accuracy and F1 score. 
Fusing the two modalities for complementation results in a noticeable improvement in F1 score, classification accuracy, and x errors. 
Since more lanes are recalled in the fusion setting, its z errors are slightly higher than LiDAR-only, which is within the expected range.

\begin{table}[!htbp]
\begin{center}
\caption{Ablation study on the number of stages of iterative regression.}
\resizebox{\linewidth}{!}{
\begin{tabular}{c|ccccc}
\toprule
\textbf{Iter} & \textbf{F1(\%)} & \textbf{E$_x$/N(m)}& \textbf{E$_x$/F(m)} & \textbf{E$_z$/N(m)} & \textbf{E$_z$/F(m)} \\
\hline 
1 & 54.9 & 0.289 & 0.296 & 0.080 & 0.110 \\
2 & 55.9 & 0.254 & 0.288 & 0.078 & 0.108 \\
3 & 56.7 & 0.253 & 0.275 & 0.078 & 0.109 \\
4 & \textbf{57.9} & \textbf{0.232} & \textbf{0.265} & \textbf{0.076} & \textbf{0.102} \\
5 & 57.6 & 0.242 & 0.262 & \textbf{0.076} & 0.103 \\
\bottomrule     
\end{tabular}}
\label{tab:iter}
\end{center}
\end{table}

\noindent \textbf{Stages of iterative regression.}
Table~\ref{tab:iter} presents the ablation results of different stages of iterative regression for Anchor3DLane++.
As the number of iteration stages increases, the initial straight-line anchor becomes increasingly aligned with the shape of the lane lines in the image, which leads to improvements in the F1 score and reductions in x/z errors.
We observe the best performance in 4 stages of iterative regression and slight performance declines in 5 stages, possibly due to overfitting caused by too many iterations.
Therefore, we choose $4$ as the number of iterative regression stages in the final implementation of Anchor3DLane++.

\begin{table}[!htbp]
\begin{center}
\caption{Ablation study on the number of sparse anchors.}
\resizebox{\linewidth}{!}{
\begin{tabular}{c|ccccc}
\toprule
\textbf{Number} & \textbf{F1(\%)} & \textbf{E$_x$/N(m)}& \textbf{E$_x$/F(m)} & \textbf{E$_z$/N(m)} & \textbf{E$_z$/F(m)} \\
\hline 
20 & 54.2 & 0.304 & 0.316 & 0.081 & 0.112 \\
30 & \textbf{54.3} & \textbf{0.295} & \textbf{0.305} & \textbf{0.079} & \textbf{0.112} \\
40 & 54.0 & 0.298 & 0.304 & 0.082 & 0.113 \\
50 & 54.0 & 0.294 & 0.313 & 0.084 & 0.115 \\
\bottomrule     
\end{tabular}}
\label{tab:anchor}
\end{center}
\end{table}

\noindent \textbf{Number of sparse anchors}.
We also conduct ablation studies on the number of sparse anchors in Table~\ref{tab:anchor}.
Increasing the number of anchors from 20 to 30 can lead to a reduction in coordinate errors. 
However, further increasing the number of anchors may result in a slight decrease in performance, possibly because too many anchors may introduce interference.
Therefore, we choose $30$ as the number of sparse anchors in the final implementation of Anchor3DLane++.

\subsection{Qualitative Results}

\noindent \textbf{Qualitative comparison on OpenLane dataset}.
We present the qualitative comparison with PersFormer on the OpenLane dataset in Fig.~\ref{fig:vis_lane}.
In the challenging conditions of steep slopes, our Anchor3DLane++ estimates the x and z coordinates more accurately as a result of direct feature sampling from FV features.
For instance, in the 1st column, the lane predicted by PersForemer noticeably deviates from the ground truth in both horizontal and vertical directions.
In the 2nd column, the angle of the left lane predicted by PersFormer is much larger than the ground truth. 
Besides, in scenarios of dense lane lines, Anchor3DLane++ results in fewer missed detections.
For example, in the 5th column, the two leftmost lanes are close to each other.
Persofmer can only predict one of them, while ours can predict both, which further proves that our sample-adaptive anchor generation approach provides better coverage of lanes on the road.

\begin{figure}[!htbp]
    \centering
    \includegraphics[width=1.0\linewidth]{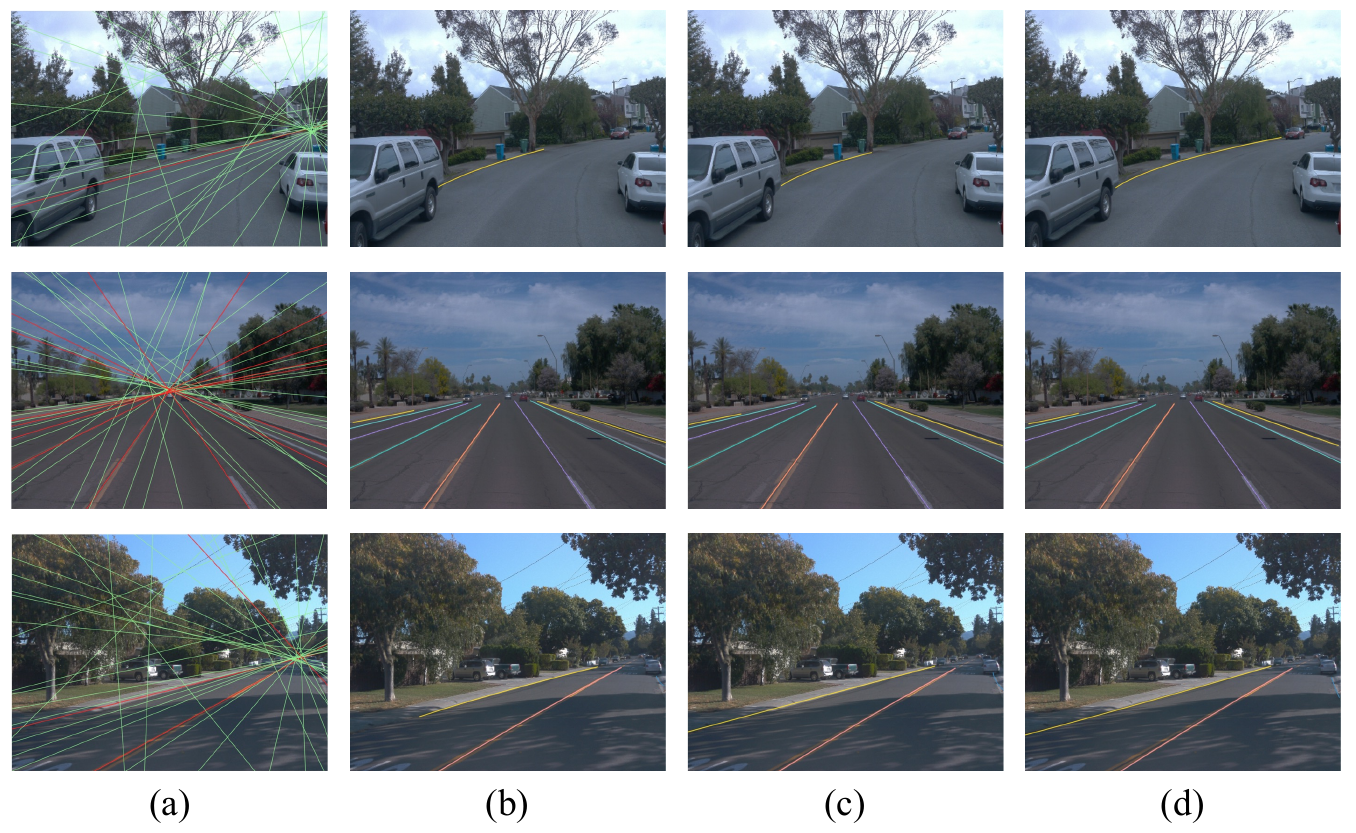}
    \caption{Visualization of initial sparse anchors and predictions of different iterative refinement stages. (a) Initial sample-adaptive anchors. Anchors that generate final predictions are drawn in red. (b) Predictions of the 1st stage. (c) Predictions of the 2nd stage. (d) Predictions of the last stage.}
    \label{fig:vis_anchor}
\end{figure}

\noindent \textbf{Visualization of intermediate results}.
To better illustrate the anchor generation and iterative refinement processes of Anchor3DLane++, we provide visualization results in Fig.~\ref{fig:vis_anchor}.
Fig.~\ref{fig:vis_anchor}(a) proves that the initial sparse anchors generated by our PAAG are well-aligned with the input images.
For example, in the 1st row, the road is oriented towards the right, and the initial anchors produced mostly follow the similar direction. 
In the second row, the ego vehicle is driving in the middle of the road, and the initial anchors are also centered in the image accordingly.
In Fig.~\ref{fig:vis_anchor}(b)-(d), it is shown that as the number of iterative stages increases, the lane predictions become gradually closer to the ground truth lanes.

\section{Conclusion}
In this work, we introduce a novel architecture named Anchor3DLane++ to detect 3D lanes from FV representations directly without introducing BEV.
3D lane anchors are designed as structural representations of 3D lanes, which are then projected to the FV features for anchor feature sampling and lane prediction.
A novel Prototype-based Adaptive Anchor Generation (PAAG) module is proposed in Anchor3DLane++ to produce sample-adaptive sparse 3D anchors dynamically.
In addition, we leverage the parallel property of 3D lanes and develop an Equal Width (EW) loss for regularization.
Moreover, we further conduct camera-LiDAR fusion based on Anchor3DLane++ to explore the benefits of information complementation.
Extensive experiments on three popular 3D lane detection benchmarks demonstrate that our Anchor3DLane++ outperforms previous state-of-the-art methods.

\textbf{Acknowledgements.} This research was supported in part by National Science and Technology Major Project (2022ZD0115502), National Natural Science Foundation of China (No. U23B2010), Zhejiang Provincial Natural Science Foundation of China (Grant No. LDT23F02022F02), Beijing Natural Science Foundation (No. L231011), and Beihang World TOP University Cooperation Program.

\bibliographystyle{IEEEtran}
\bibliography{IEEEabrv,ref}

\begin{IEEEbiography}[{\includegraphics[width=1in,height=1.25in,clip,keepaspectratio]{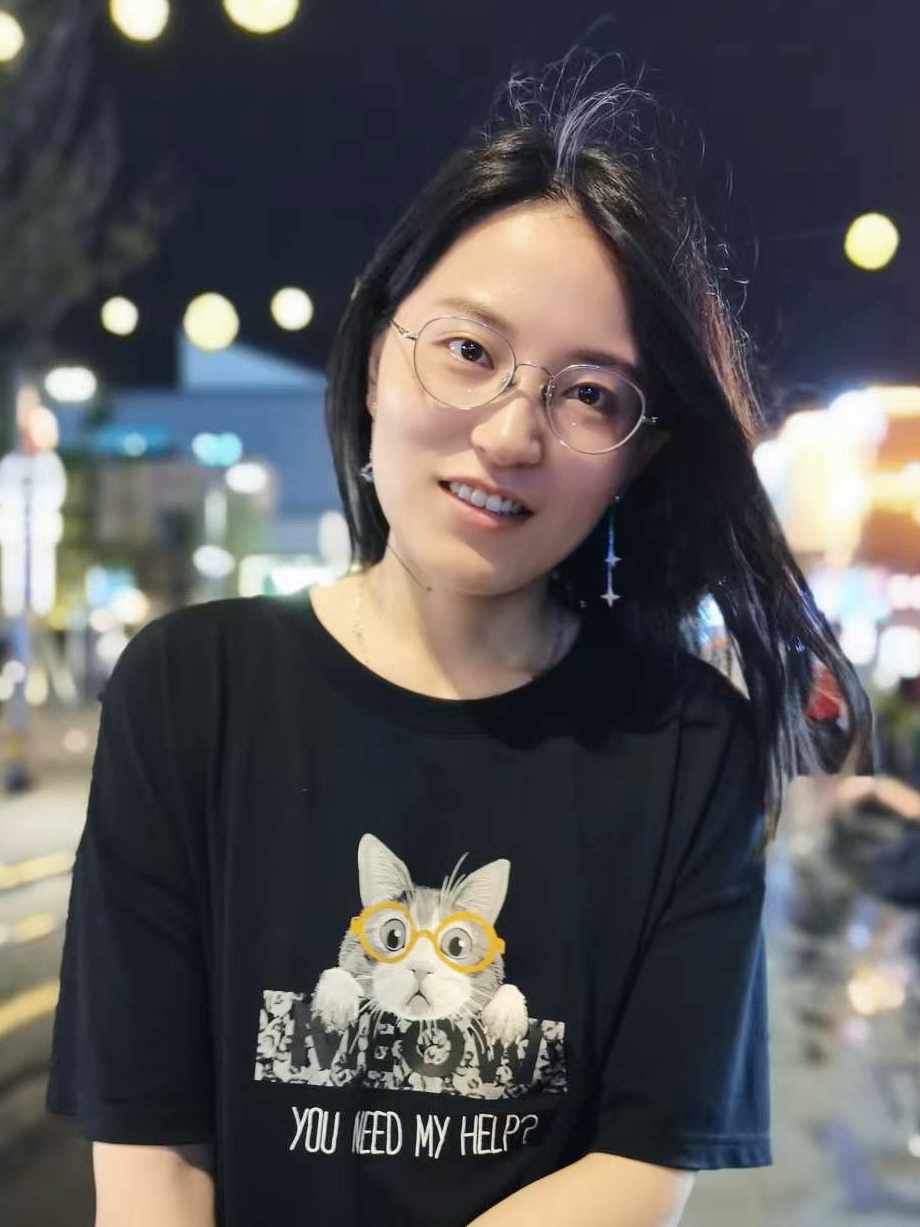}}]{Shaofei Huang}
is currently a Ph.D. candidate at Institute of Information Engineering, Chinese Academy of Sciences, supervised by Prof. Si Liu and Prof. Jizhong Han. She received her B.S. degree from Peking University. She has published several papers on CVPR, ECCV, TIP, etc. Her research interests include image and video segmentation, and road element perception for autonomous driving.
\end{IEEEbiography}

\begin{IEEEbiography}[{\includegraphics[width=1in,height=1.25in,clip,keepaspectratio]{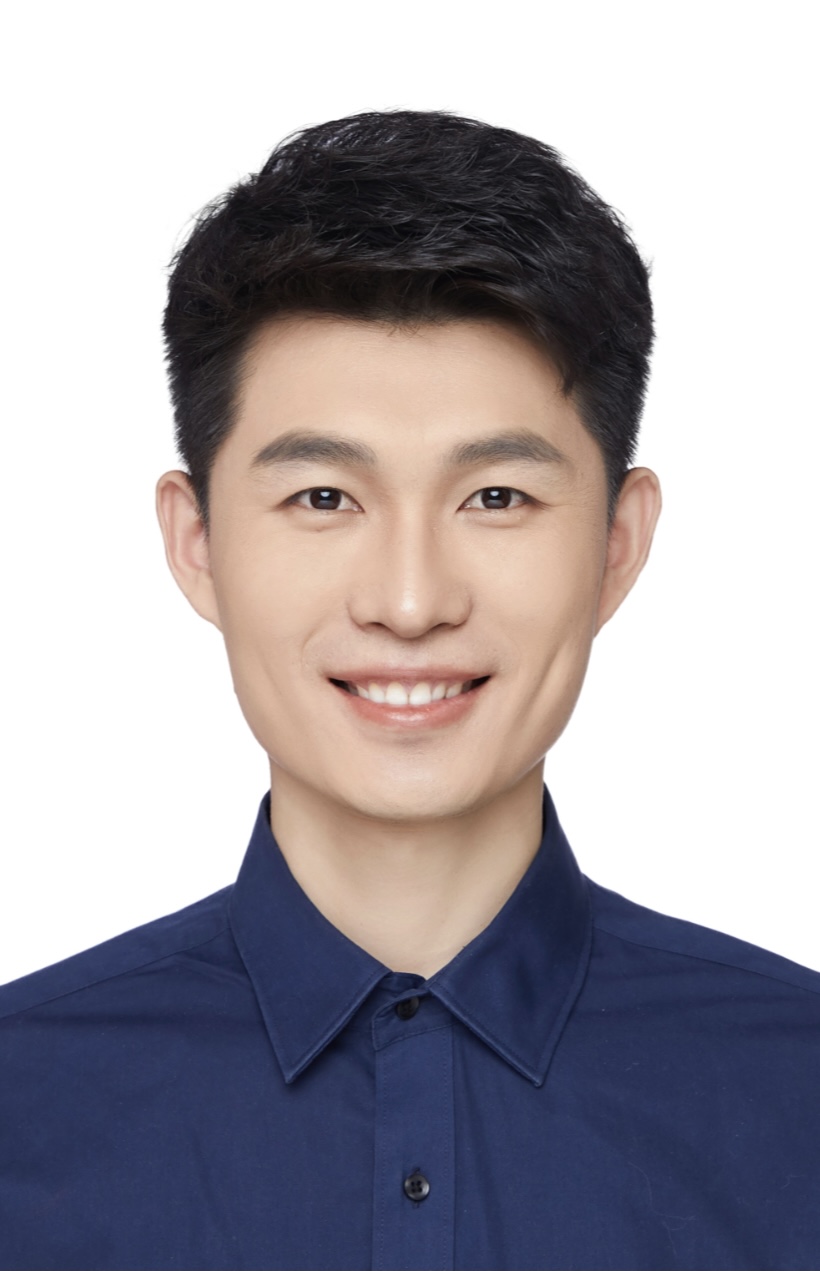}}]{Zhenwei Shen} is currently an algorithm engineer at TuSimple. He received his B.S. degree in Vehicle Engineering from Shandong University of Science and Technology, Qingdao, China, in 2014, and his M.E. degree in Transportation Engineering from Shandong University of Science and Technology, Qingdao, China, in 2018. His research interests include computer vision and deep learning.
\end{IEEEbiography}

\begin{IEEEbiography}[{\includegraphics[width=1in,height=1.25in,clip,keepaspectratio]{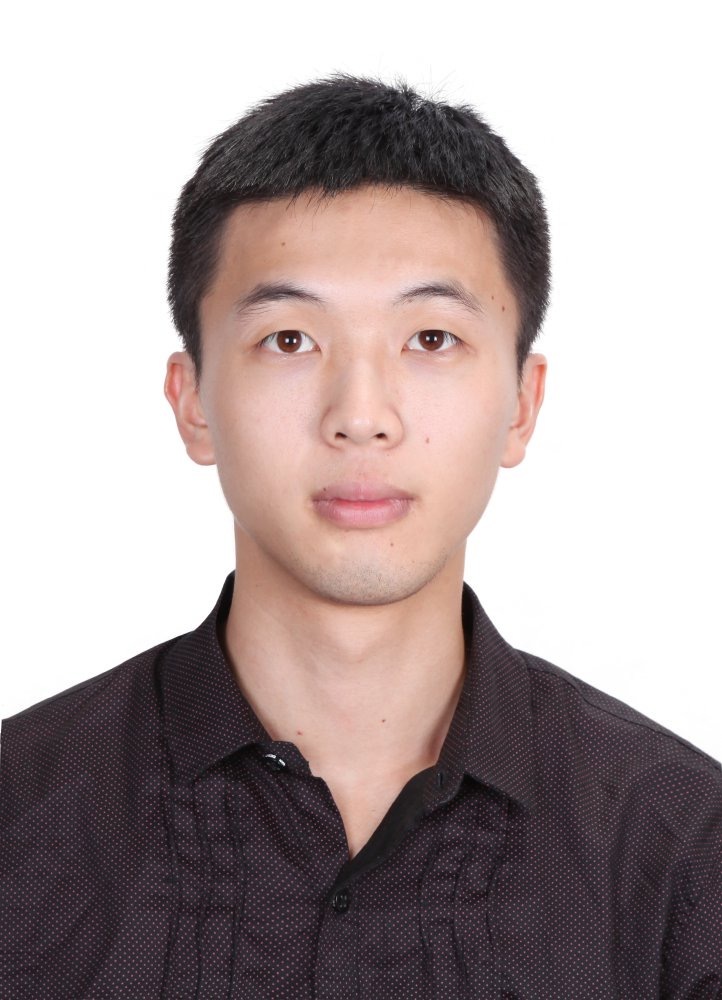}}]{Zehao Huang}
is currently an algorithm engineer at TuSimple. He received his B.S. degree in automatic control from Beihang University, Beijing, China, in 2015. His research interests include computer vision and image processing.
\end{IEEEbiography}

\begin{IEEEbiography}[{\includegraphics[width=1in,height=1.25in,clip,keepaspectratio]{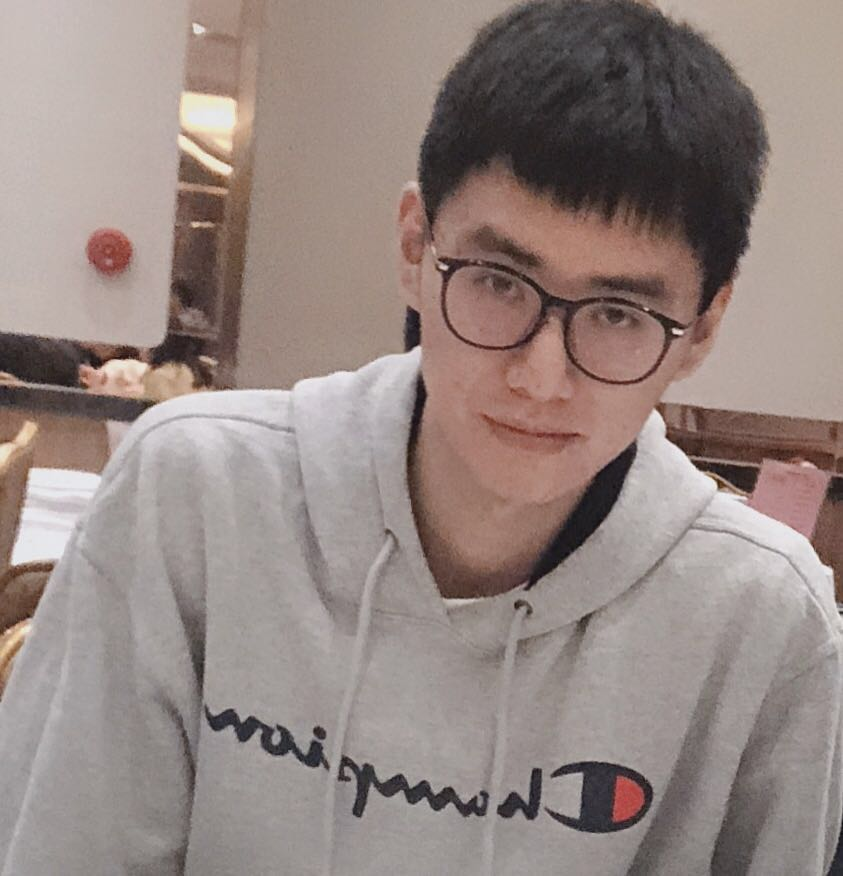}}]{Yue Liao} is currently a post-doctoral fellow in Multi-Media Lab (MMLab) at The Chinese University of Hong Kong (CUHK) and Centre for Perceptual and Interactive Intelligence (CPII). He received his PhD degree at School of Computer Science and Engineering, Beihang University.  His research interests include human-object interaction detection, multi-modality understanding and object detection. He has published more than 10 papers at top journals and conferences, including T-PAMI, T-IP, NIPS, CVPR, ICCV and ECCV, etc.
\end{IEEEbiography}

\begin{IEEEbiography}[{\includegraphics[width=1in,height=1.25in,clip,keepaspectratio]{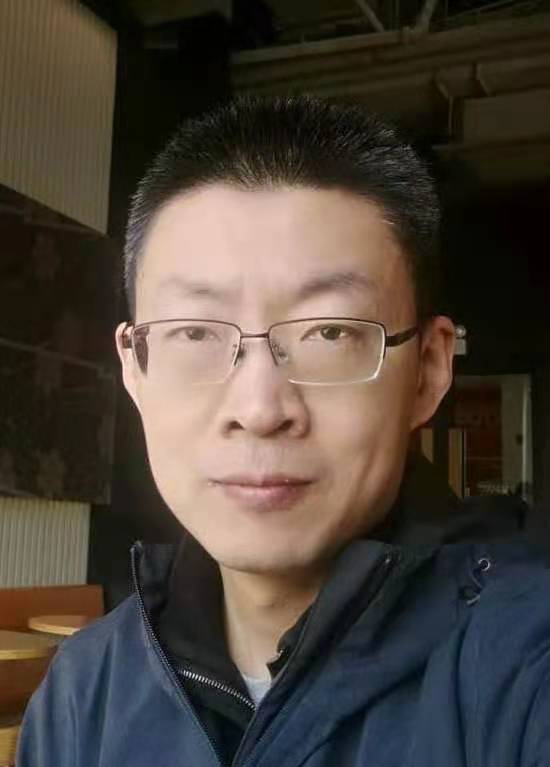}}]{Jizhong Han}
is currently a full professor at Institute of Information Engineering, Chinese Academy of Sciences. He received his Ph.D. degree from Institute of Computing Technology, Chinese Academy of Sciences. His research interests include multimedia information processing and big data storage. He has published over 60 papers and held over 10 domestic patents. He is also the principal investigator or participant of several National 973 or 863 Programs.
\end{IEEEbiography}

\begin{IEEEbiography}[{\includegraphics[width=1in,height=1.25in,clip,keepaspectratio]{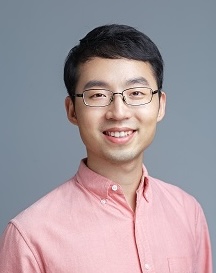}}]{Naiyan Wang}
is currently the chief scientist of TuSimple and leads the algorithm research group in the Beijing branch. Before this, he got his Ph.D. degree from the CSE department, Hong Kong University of Science and Technology in 2015. His supervisor is Prof. Dit-Yan Yeung. He got his B.S. degree from Zhejiang University, in 2011 under the supervision of Prof. Zhihua Zhang. His research interest focuses on applying the statistical computational model to real problems in computer vision and data mining. Currently, he mainly works on the vision-based perception and localization part of autonomous driving. Especially he integrates and improves the cutting-edge technologies in academia, and makes them work properly in the autonomous truck.
\end{IEEEbiography}

\begin{IEEEbiography}[{\includegraphics[width=1in,height=1.25in,clip,keepaspectratio]{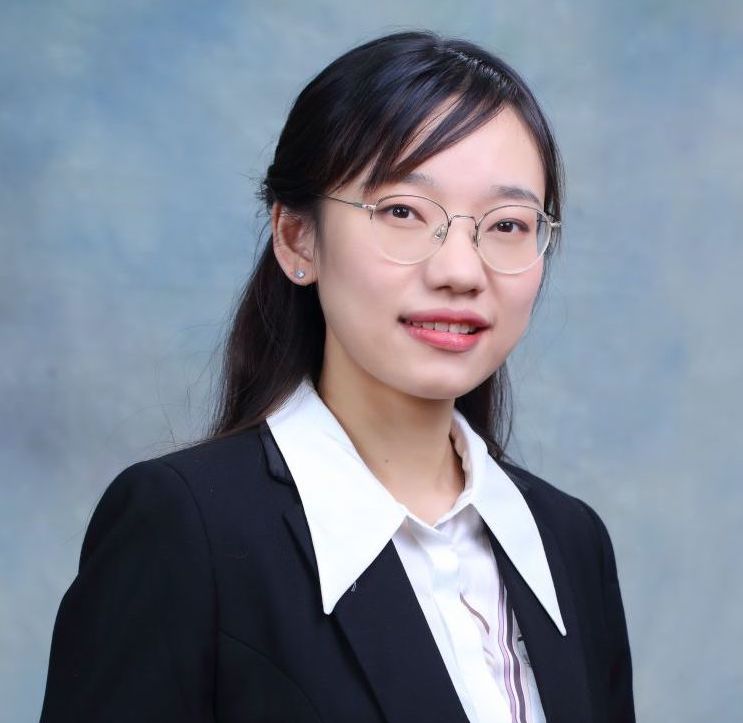}}]{Si Liu}
is currently a full professor at Institute of Artificial Intelligence, Beihang University. She is the recipient of the National Science Fund for Excellent Young Scholars. Her research interests include cross-modal multimedia intelligent analysis and classical computer vision tasks. She has published more than 60 cutting-edge papers and been cited over 13000 times on Google Scholar. She has won the Best Paper Awards of ACM MM 2021 and 2013, the Best Video Award of IJCAI 2021, and the Best Demo Award of ACM MM 2012. She is currently the associate editor of IEEE TMM, IEEE TCSVT, and CVIU, and she has served as the area chair of ICCV, CVPR, ECCV, ACM MM, and other top conferences many times.
\end{IEEEbiography}

\vfill

\end{document}